# Large language models reshape the language of science

Dingkang Lin[*], Naixuan Zhao[*], Dan Tian[**] and Jiang Li [*]

[*]*652023140004@smail.nju.edu.cn; 502024140049@smail.nju.edu.cn; lijiang@nju.edu.cn*
0000-0001-7275-8118; 0009-0008-6649-5129; 0000-0001-5769-8647
School of Information Management, Nanjing University, Nanjing, China, 210032

[**] *dan_tian@sxu.edu.cn*
0000-0003-1980-8467
School of Economics and Management, Shanxi University, Taiyuan, China, 030031

# Abstract

Scientific language is a central infrastructure of knowledge production, but it remains unclear whether large language models (LLMs) are altering not only how scientists write, but also how scientific knowledge is communicated and accessed. Here we analyze 21.36 million scientific abstracts published between 2020 and 2024, together with historical records from major journals, to trace recent changes in the language of science. We identify a marked turning point in 2024, when scientific writing shows a sharp increase in lexical complexity alongside a decline in syntactic complexity. This shift is pervasive across disciplines and journal tiers, and is more pronounced in texts by scholars working in non-native English contexts, especially those from language backgrounds that differ more typologically from English. Controlled polishing experiments confirm that LLMs reproduce this pattern by favoring more lexically dense and syntactically compressed expression. We further show why this linguistic shift matters: it may widen the distance between scientific discourse and public-facing language, while also helping scholars in non-native English contexts navigate English-language publishing requirements. These findings suggest that LLMs may broaden participation in scientific authorship while narrowing the accessibility of scientific communication, making them a new force in the linguistic infrastructure of science.

# Introduction

Scientific language is not merely a style of writing but a central infrastructure of knowledge production. It organizes how findings are formulated, how evidence is evaluated, how expertise travels across communities, and how research enters global circulation. In contemporary science, this infrastructure is overwhelmingly mediated by English: English dominates indexed scientific publishing and functions as the principal medium through which scholars across linguistic backgrounds communicate, compete, and gain recognition (Liu, 2017; Kourilova-Urbanczik, 2012). Changes in scientific English therefore have consequences beyond wording. They can alter the form in which knowledge is compressed, transmitted, and made accessible (Higham & Nagaoka, 2026).

Large language models (LLMs) have introduced a new force into this linguistic infrastructure (Shin et al., 2026). Since the public release of ChatGPT in November 2022, LLMs have been rapidly incorporated into academic workflows, including text composition (Kobak et al., 2025; Liang et al., 2025), literature review writing (Arif et al., 2023; Lieberum et al., 2025), experimental design and data processing (Akata et al., 2025; Bucaioni et al., 2024; Burger et al., 2023), and hypothesis generation and result analysis (Park et al., 2024; Luo et al., 2025). Their influence is especially direct in scientific writing, where they do not simply assist authors but generate, translate, and polish the linguistic forms through which scientific claims are expressed. Recent estimates suggest that LLM-influenced writing has become widespread in scholarly texts (He & Bu, 2026), exceeding 13% of preprints by late 2024 and surpassing 20% in fields such as computer science (Kobak et al., 2025; Liang et al., 2025).

Existing research has primarily documented the presence of LLMs in science rather than their structural consequences for scientific language. Some studies infer LLM use through detection tools or through the rising frequency of characteristic words and phrases (Akram, 2024; Berdejo-Espinola & Amano, 2023; Kobak et al., 2024; Kobak et al., 2025; Liang et al., 2024), while others show that adoption varies across fields, venues, and author groups (Cheng et al., 2024; Liang et al., 2024; Picazo-Sanchez & Ortiz-Martin, 2024). A related literature has examined linguistic differences between native English-speaking (NES) and non-native English-speaking (NNES) scholars (Bao et al., 2025; Liu & Bu, 2024). Yet these studies leave open a broader question: are LLMs merely leaving detectable traces in academic prose, or are they changing the structure of scientific language itself?

This question matters because scientific English has long followed its own historical trajectory. Prior work shows that academic discourse has moved toward greater information density, nominalization, and phrasal compression, replacing clausal elaboration with compact noun-centered structures (Biber & Gray, 2010; Halliday & Martin, 1993). Historical disruptions, including wars and technological shifts, have also altered scientific vocabularies and patterns of expression (Bizzoni et al., 2020; Bochkarev et al., 2014). As our historical data show, lexical complexity in leading scientific journals has risen steadily since 1950, while syntactic complexity peaked in the late twentieth century and subsequently declined (Fig. 1A). The emergence of LLMs raises the possibility that this long-term evolution is now being accelerated by a technology that operates directly on language.

Here we examine whether LLMs are reshaping the language of science at scale, and why such a shift matters for science itself. We combine a large-scale corpus analysis of 21.36 million scientific abstracts from 2020 to 2024 with historical comparisons, quasi-experimental designs comparing NES and NNES author groups, and controlled LLM polishing experiments that measure how texts change before and after model-assisted editing. Across these sources of evidence, we find a clear post-2023 shift: scientific English becomes more lexically complex and information-dense while becoming syntactically simpler and more compressed. We then connect this linguistic change to two scientific consequences: the accessibility of scientific communication and the participation of scholars who face additional English-language writing demands. This framing moves the study beyond detecting LLM traces in prose to asking how LLMs may change the communicative and participatory infrastructure of science.

# Measurement framework

To quantify the complexity of academic language, we employ a multidimensional framework assessing lexical and syntactic dimensions. For lexical complexity, we measure lexical diversity via MTLD (McCarthy & Jarvis, 2010), which stabilizes type-token ratio (TTR) at 0.72 to mitigate text-length bias. We measure lexical sophistication through Word Length (mean characters per word) and Syllables per Word (mean syllables per word). Lexical density is measured by Noun Ratio, Adjective Ratio, Adverb Ratio, and Verb Ratio (Lu, 2012), representing proportions of content words. For syntactic complexity, we adopted Lu's (Lu, 2010; Lu, 2011) 14-indicator system spanning five dimensions: Production Unit Length for clause, sentence and T-unit; Sentence Complexity for clause embedding density; Subordination for clause and T-unit complexity; Coordination for phrase-level and sentence-level parallelism; Specific Structures for nominal and verbal complexity. The indicator system is summarized in Supplementary Table S1.

These metrics represent widely adopted measures for assessing both syntactic and lexical complexity in academic writing. By computing all indicators comprehensively, our approach enables cross-validation of results while capturing granular shifts in linguistic complexity across distinct dimensions. Our selection criteria prioritize three key attributes: field ubiquity, strong interpretability, and multidimensional coverage. Given metric multiplicity, we aggregated indicators into lexical/syntactic composite indices for visualization in Figure 1. Our rationale for metric selection, full computational formulae, definitions, and linguistic interpretations appear in Appendix S1.

# Results

## *Scientific language has become more compressed in the LLM era*

We first examined whether recent scientific English shows evidence of a system-level linguistic shift rather than isolated stylistic fluctuation. Using the OpenAlex database, we retrieved journal articles and conference proceedings published between January 1, 2020, and November 1, 2024. After filtering for English-language publications with valid abstracts and at least one author, our core corpus comprised 21.36 million abstracts from 2020 to 2024. To place these recent changes in historical context, we supplemented the corpus with 375,000 abstracts from Nature, Science, and Proceedings of the National Academy of Sciences of the United States (PNAS) spanning 1950 to 2020. Author ethnicity was classified via Ethnicseer (85% accuracy) (Treeratpituk & Giles, 2021), while country affiliation was determined from institutional metadata. Full data processing details appear in Appendices S2–S3.

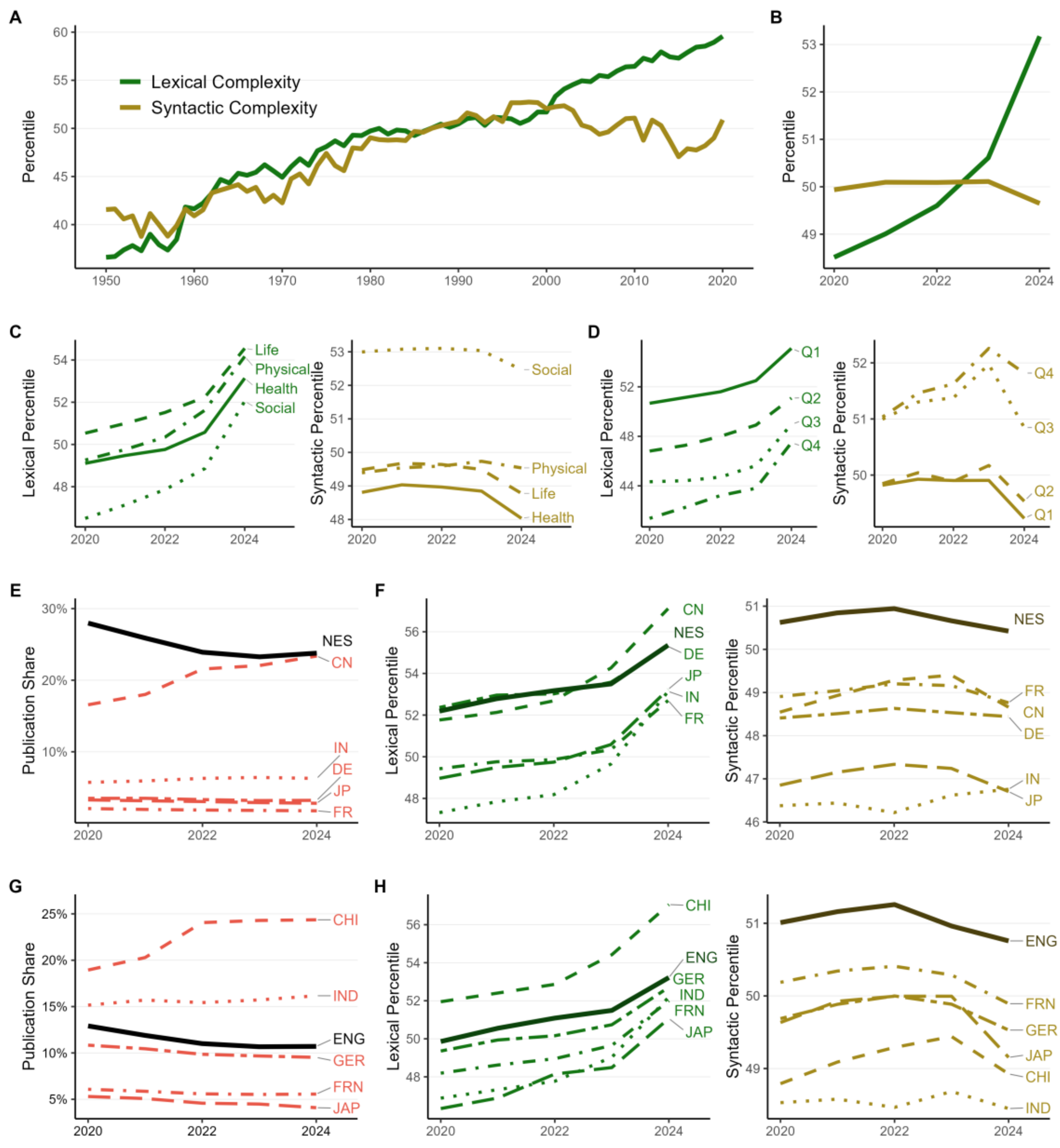

Figure 1. Linguistic compression and author-group shifts in scientific abstracts. (A) Percentile trends of mean linguistic complexity in Nature, Science, and PNAS abstracts (1950–2020). (B) Complexity evolution across 21.36 million article abstracts during 2020–2024. (C) Domain-stratified trends (Life Sciences, Physical Sciences, Health Sciences, Social Sciences) during 2020–2024. (D) Journal Impact Factor (JIF) quartile-stratified trends (Q1–Q4) during 2020–2024. (E) Contribution share to scientific publications by country of institutional affiliation (2020–2024), with NES indicating contributions from primary English-speaking countries. (F) Country-level changes in mean linguistic complexity during 2020–2024. (G) Contribution share by author ethnicity during 2020–2024. (H) Ethnicity-based changes in mean linguistic complexity during 2020–2024.

Long-run trends show that scientific English has been moving toward increasingly compressed expression. From 1950 to 1998, linguistic complexity increased across both lexical and syntactic dimensions, after which syntactic complexity plateaued and gradually declined (Fig. 1A). As shown in Appendix Fig. S1, the rise in lexical complexity primarily reflects increasing word length and lexical density, indicating greater use of sophisticated terminology and content words relative to function words. The post-1990s decline in syntactic complexity (Appendix Fig. S2) reflects shorter production units and reduced use of complex nominals; previously increasing features such as clauses per sentence, subordination, and coordination also ceased growing during this period.

A pronounced turning point occurred in 2024: lexical complexity surged while syntactic complexity declined (Fig. 1B). This pattern suggests a shift toward more information-dense and compressed scientific language rather than a simple increase or decrease in overall complexity. Appendix Fig. S3 shows that lexical intensification spans nearly all dimensions, including word length, lexical diversity, and density metrics, with the exception of adverb ratio. At the same time, syntactic simplification appears through shorter production units and reduced use of dependent clauses and complex nominals. Scientific abstracts thus became more terminologically dense while relying on less elaborate sentence architecture.

### *The shift is pervasive across fields and journal tiers*

The 2024 shift was not confined to a particular field, venue type, or quality tier, indicating a broader change in the language of science. Using OpenAlex's domain classification (field mappings in Table S2), we found that every major research domain exhibited steeper lexical increases and syntactic declines in 2024 compared with 2020–2023 trends (Fig. 1C). This pattern also held across all 26 subfields (Appendix Fig. S4). Journal Impact Factor quartile analysis (methodology in Appendix S2.1) showed the same direction of change across Q1–Q4 journals (Fig. 1D). Social sciences showed the lowest lexical complexity but the highest syntactic complexity, while higher-tier journals (Q1/Q2) displayed higher lexical sophistication alongside lower syntactic intricacy. These cross-field and cross-tier patterns suggest that LLM-era language change reflects a system-wide transformation of scientific writing rather than a local anomaly.

### *The transformation is amplified among non-native English-speaking scholars*

If LLMs are reshaping scientific English, the transformation may vary across author groups. Scientific writing is embedded in a global English-language system in which scholars writing outside their first language often face additional language-related demands. We therefore examined whether the 2024 compression pattern was more pronounced among NNES authors, who may have greater incentives to use LLM-based translation, drafting, and polishing tools.

Country-level patterns support this expectation. Figure 1E shows a declining relative publication share from predominantly English-speaking countries (NES: United States, United Kingdom, and related countries) alongside rising contributions from several NNES countries, notably China. Figure 1F shows that authors affiliated with NNES countries, especially countries whose dominant languages differ more typologically from English, such as China and Japan, underwent larger 2024 shifts in both lexical and syntactic dimensions than NES authors. Countries whose dominant languages are more typologically similar to English, such as Germany and France, showed trajectories more similar to NES groups.

Ethnicity-stratified analyses corroborate these patterns (Fig. 1G–H). Publications by English-ethnic authors declined in relative share, while Chinese- and Indian-ethnic authorship increased. Mirroring the country-level results, Japanese and Chinese author groups showed stronger 2024 shifts than English-ethnic authors across both lexical and syntactic measures. These descriptive patterns indicate that LLM-era changes in scientific language are shaped by asymmetries in the global use of English in science.

However, these descriptive analyses have important limitations. Complexity metrics weighted by author contribution can blend linguistic features from multi-author teams, institutional affiliation may not capture actual language background because of global academic mobility, and name-based ethnicity inference cannot measure fluency or name changes. We therefore used more stringent quasi-experimental designs to test whether NNES texts diverged from NES texts after widespread LLM adoption.

To isolate the post-ChatGPT divergence between NNES and NES scientific writing, we implemented a difference-in-differences framework.
We defined treatment and control groups through stringent criteria (Fig. 5): the NNES group comprised articles exclusively authored by name-inferred Chinese researchers affiliated with Chinese institutions (2.47 million articles), while the NES control group comprised publications solely by name-inferred English researchers at U.S./U.K. institutions (331,000 articles). This screening strategy maximized group consistency. Our dual-level approach incorporated journal-level analyses controlling for year and venue fixed effects and author-level analyses controlling for year and first-author fixed effects, enabling comparisons within identical publication contexts and across individual writing trajectories.

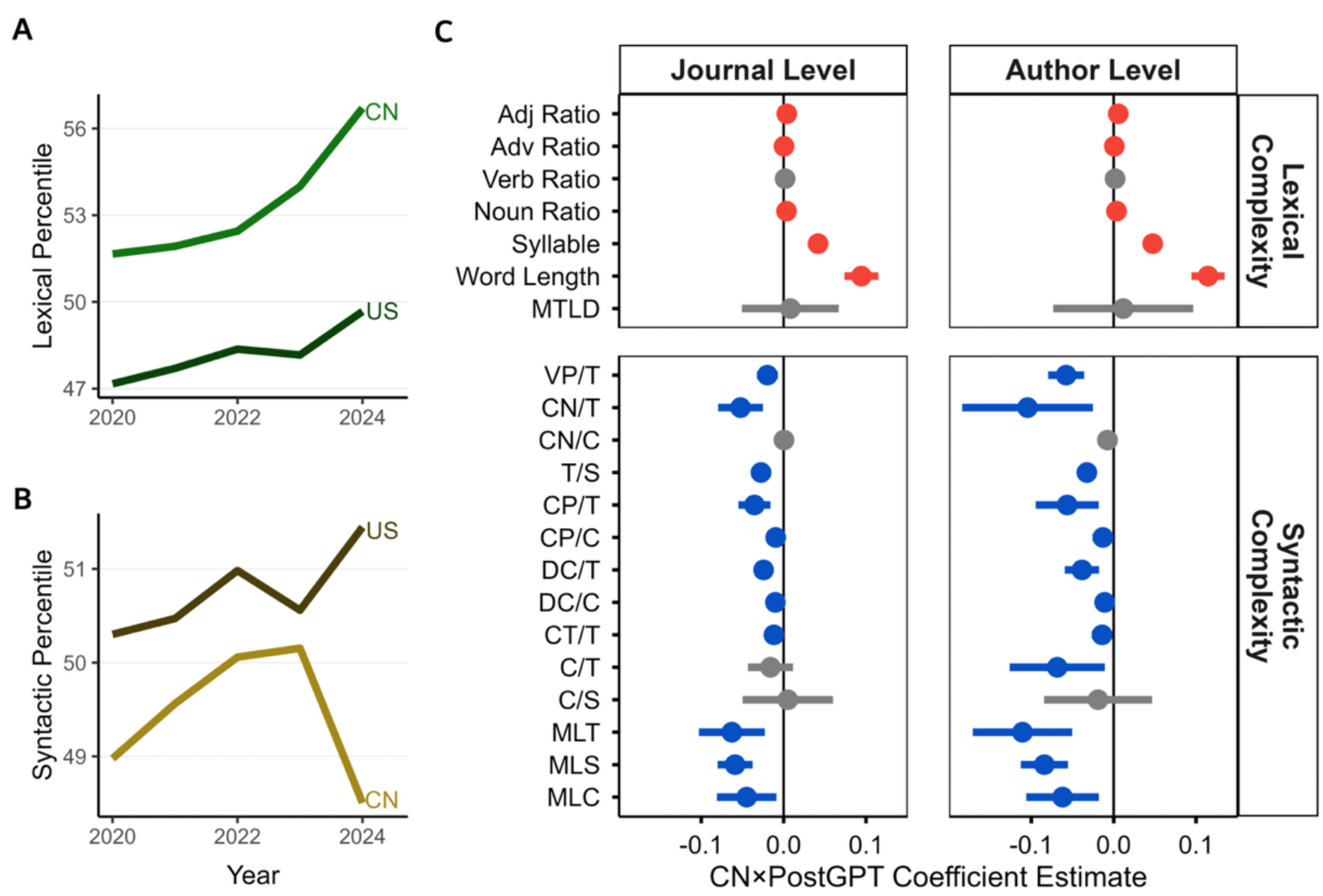

Figure 2. Post-ChatGPT divergence in compressed scientific English between NNES and NES authors. (A) Lexical complexity trajectories (mean percentiles) for NNES and NES authors during 2020–2024. (B) Syntactic complexity trajectories (mean percentiles) for corresponding groups. (C) Coefficient estimates for the CN×PostGPT interaction term across all 21 complexity metrics (rows) and analytical levels (columns). CN is the treatment group indicator,

PostGPT = 2024. Red indicates positive coefficients with $p<0.05$ significance, blue indicates negative coefficients with $p<0.05$ significance, and gray indicates statistically non-significant estimates.

Figure 2A and 2B reveal a pronounced 2024 divergence: texts in the Chinese NNES sample became more lexically complex while becoming syntactically simpler relative to NES controls. Appendix S11 and Fig. S5 show direct mean values across specific indicators, demonstrating clear 2024 deviations for most metrics in the Chinese sample. The U.S./U.K. NES control group shows some mean-value fluctuation, which our data audit did not attribute to anomalies; instead, it reflects the smaller sample generated by strict control-group construction in a highly mobile and collaborative academic system. Because NES authors may also use LLMs, these estimates should be interpreted as conservative lower bounds of differential LLM influence.

The regression evidence indicates that the NNES transformation is not merely a descriptive artifact. Lexical density increased significantly for adjectives, verbs, and nouns, while adverb ratios remained largely unchanged. Lexical sophistication also increased, particularly word length, which rose by an average of 0.1 characters per word; lexical diversity showed no significant LLM-driven change. Syntactically, sentence and clause lengths declined, and the use of dependent clauses, coordinate phrases, and complex nominals decreased (Fig. 2C; Fig. S7). These changes amount to a more compressed mode of scientific expression: denser lexical packaging with less clausal elaboration.

Additional analyses support this interpretation. Fig. S6 shows no significant changes in basic constituent counts, such as clause counts, sentence counts, or verb phrases, indicating that the observed syntactic shifts reflect structural streamlining rather than isolated changes in individual elements. Extending the framework to other author groups (Fig. S8) reveals a gradient associated with typological distance from English: Japanese and Korean author groups show effects similar to those observed in the Chinese NNES sample, German and French author groups show minimal or inconsistent changes, and Indian and Italian author groups exhibit intermediate patterns.

Domain-stratified analyses confirm that this amplified NNES transformation occurs within the broader system-wide shift (Figs. S9–S10). All Journal Impact Factor quartiles show the core pattern, with stronger statistical significance in Q1–Q2 journals, and all major research domains exhibit characteristic changes, although effect sizes vary. Social sciences show the strongest gains in lexical sophistication. Together, these results show that LLM-era scientific language change is pervasive, but unevenly distributed across the global English-language structure of science.

### *LLM polishing reproduces the observed linguistic shift*

The preceding analyses provide converging evidence that scientific English shifted in 2024 and that the shift was amplified among NNES scholars, but observational designs cannot by themselves establish a complete causal chain. We therefore conducted controlled text-polishing experiments to test whether LLM editing directly reproduces the same pattern of lexical densification and syntactic compression.

We randomly selected 1,000 abstracts authored by Chinese NNES researchers in 2020, before widespread ChatGPT adoption. These abstracts were processed through ChatGPT-4o for polishing, and post-polish complexity metrics were compared with the originals using paired t-tests. To assess robustness, we replicated the experiment with alternative LLMs (DeepSeek, LLaMA, Qianwen), ChatGPT versions (v3.5, v4, v4o), distinct prompt formulations, and supplementary samples. Full experimental protocols appear in Appendix S5.

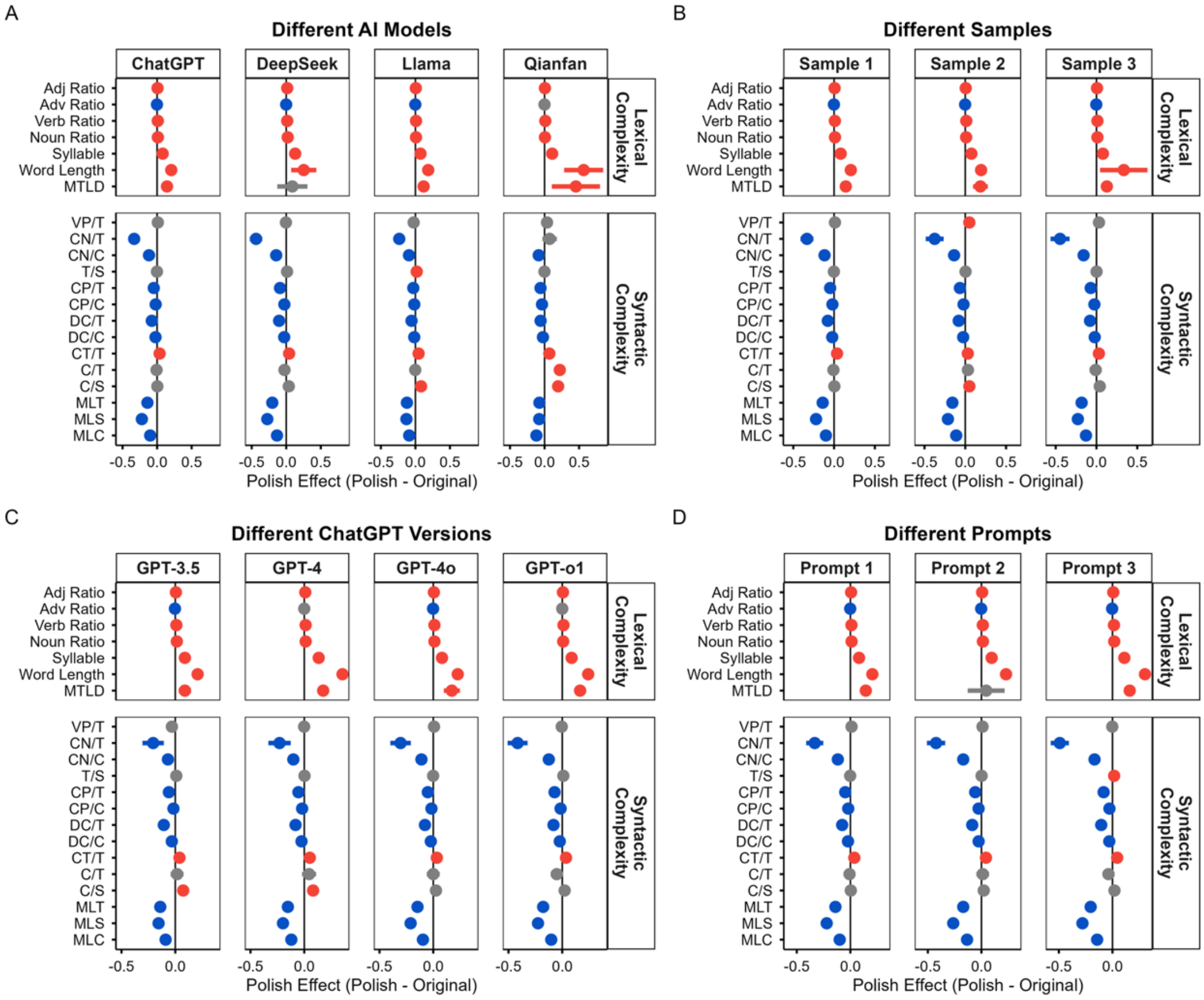

Figure 3. Experimental validation of LLM-induced linguistic compression. Randomly sampled abstracts from the Chinese NNES sample were polished by large language models, after which linguistic complexity changes were quantified using paired t-tests. (A) Cross-model comparison across ChatGPT, DeepSeek, LLaMA, and Qianwen. (B) Cross-sample generalization using ChatGPT. (C) Version-specific effects across ChatGPT iterations: v3.5, v4, v4o, and v-o1. (D) Prompt-engineering impacts using three instructional prompts. Red indicates positive coefficients with $p<0.05$ significance, blue indicates negative coefficients with $p<0.05$ significance, and gray indicates statistically non-significant estimates.

The experiments reproduced the observational pattern. Across models, versions, prompts, and sample selections, LLM polishing consistently increased lexical sophistication while streamlining syntactic structures (Fig. 3). This stability across experimental conditions shows that the observed 2024 pattern is not simply a byproduct of field composition, publication venue, or time trend.

Rather, it reflects a property of LLM-mediated rewriting itself: models tend to favor denser word choice and more compact syntactic packaging.

Parallel processing of NES-authored abstracts (Fig. S11) confirmed the same directional shift: lexical complexity increased and syntactic complexity decreased after polishing. This indicates that LLM-induced compression is not restricted to NNES texts. However, effect-size analyses show that the magnitude of transformation differs across author groups.

Appendix S12 compares texts from the Chinese NNES and English NES samples before and after polishing. Two patterns emerge. First, within texts from the same author group, polishing effects are stable across models, prompts, and samples. Second, cross-group differences in baseline complexity remain highly persistent after polishing. Thus, LLMs induce a common direction of language change, but they operate on pre-existing differences in writing patterns rather than erasing them. The result is a shared movement toward compressed scientific English whose magnitude varies across author groups.

### *LLM-driven language change affects scientific communication and participation*

Finally, we examined why LLM-driven linguistic change matters beyond text-level metrics. If scientific English is becoming more compressed, the consequences should be visible in two domains central to science: how far scientific discourse moves away from public-facing language, and whether LLM-assisted writing changes participation in an English-dominated publication system.

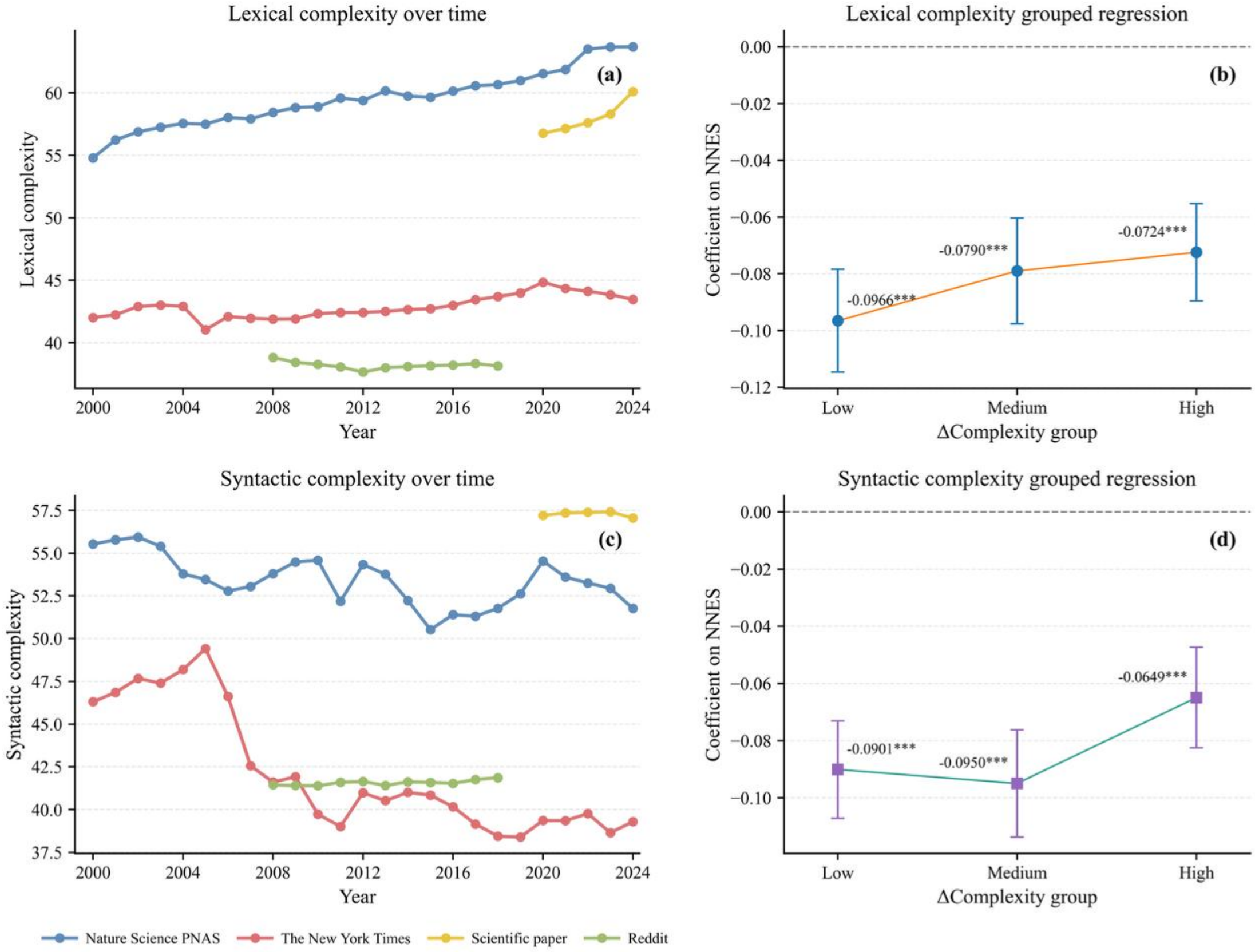

Figure 4. Historical Trends in Linguistic Complexity and Its Association with Author Productivity. (a) Lexical complexity trajectories (2000-2024) for four text categories: scientific paper abstracts (paper), abstracts from high-impact journals (Nature, Science, PNAS), New York Times article summaries, and Reddit forum text. The percentile rank for each text is calculated relative to the entire pooled corpus (all texts from all categories and years), positioning all registers on a common scale for direct comparison. (c) Syntactic complexity trajectories (2000–2024). (b) Grouped regression coefficients illustrating the differential effect of lexical complexity change on publication output change for NNES authors relative to NES authors. Authors were grouped into bins based on the magnitude of their change in textual lexical complexity (x-axis). For each bin, a coefficient (y-axis) was estimated from a regression where the dependent variable is the change in publication count, and the key independent variable is an interaction between NNES author status and the bin indicator. (d) Grouped regression coefficients derived from the same method as in (b), but using the change in syntactic complexity for grouping authors on the x-axis.

Cross-register comparisons place scientific abstracts, high-impact journal abstracts, New York Times summaries, and Reddit conversations on a common complexity continuum. Scientific abstracts occupy the highest percentiles in lexical complexity, with a marked 2024 increase that widens their distance from news and online conversational discourse (Fig. 4A). Their syntactic complexity remains higher than non-specialist texts but shows a downward trajectory (Fig. 4C). High-impact journal abstracts exhibit the most advanced form of this compression: greater lexical sophistication combined with lower syntactic complexity. These patterns indicate a substantive communication consequence: LLMs may make scientific writing more publication-like and information-dense while also making it less transparent to non-specialist and cross-disciplinary readers.

At the same time, the shift may affect participation in English-language science. We related authors' changes in linguistic complexity to changes in publication output and found that increases in lexical complexity among NNES authors were associated with a relative improvement in annual publication output compared with NES authors (Table S13; Fig. 4B). This association suggests a second substantive consequence: LLM-assisted writing may help reduce some language-related barriers faced by scholars writing in a non-native language. Taken together, the results show that LLM-driven language change is not merely stylistic. It may broaden participation in scientific authorship, even as it may narrow the accessibility of scientific communication.

# Discussion

Our findings show that large language models (LLMs) are not merely adding detectable traces to scientific prose; they are reshaping the linguistic form through which scientific knowledge is communicated, evaluated, and accessed. The central contribution is therefore not only that scientific language changed in 2024, but that this change has consequences for the organization of scientific communication. By pushing scientific English toward denser lexical packaging and reduced syntactic elaboration, LLMs may alter what is recognized as polished scientific expression, who can more easily produce it, and who can readily understand it.

### *LLMs as a force in the evolution of scientific language*

Scientific English has never been static. Prior work has shown that academic discourse has long moved toward greater informational density, nominalization, and phrasal compression, replacing clausal elaboration with compact noun-centered structures (Biber & Gray, 2010; Halliday & Martin, 1993). Our results suggest that LLMs accelerate this historical trajectory. The distinctive 2024 pattern—increased lexical complexity alongside reduced syntactic complexity—does not represent a random stylistic fluctuation. It reflects a systematic movement toward scientific prose that packages more information into denser lexical units while relying on simpler sentence architecture.

This matters because LLMs intervene directly at the level of language production. Earlier forces shaping scientific English, such as disciplinary specialization, journal norms, and the globalization of English, typically operated through slow institutional and communicative pressures. LLMs act differently: they are embedded in everyday drafting, translation, and polishing practices, and they can rapidly standardize preferred forms of expression across fields. In this sense, LLMs function as linguistic infrastructure. They do not simply help authors express pre-existing ideas more fluently; they influence the formal patterns through which scientific claims are made readable, publishable, and recognizable as academic English.

### *Why the shift differs between NES and NNES scholars*

The amplified shift among NNES scholars likely reflects a combination of adoption intensity and pre-existing linguistic structure. One possibility is differential use: NNES scholars may rely more heavily on LLMs for translation, drafting, and sentence-level polishing than NES scholars. We attempted to assess this mechanism using AIGC detection tools, but these tools showed limited reliability in scientific texts (Appendix S6.3), preventing a direct behavioral estimate of adoption intensity. A second possibility is baseline proximity: NES writing might already resemble LLM-preferred scientific English and therefore change less after polishing. Our controlled experiments do not support this as a complete explanation, because NES-authored texts also undergo the same directional shift when processed by LLMs.

A more plausible interpretation is that LLMs operate on top of stable ethnolinguistic differences in baseline writing patterns. Regression analyses show persistent cross-ethnic differences in linguistic complexity signatures (Appendix S6.2; Fig. S16), and polishing experiments show that LLMs modulate but do not erase those differences (Fig. S13; Appendix S12). Thus, LLMs appear to induce a common movement toward compressed scientific English while preserving differences rooted in language background and writing habits. Behavioral variation may further amplify this pattern: some authors may use LLMs only for light editing, whereas others may use them for translation, rewriting, or full drafting. Future work should directly measure these usage modes, but the current evidence suggests that LLM-driven language change is filtered through the unequal linguistic conditions of global science.

### *Consequences for scientific communication*

The compression of scientific English has a double edge. On one hand, denser terminology and streamlined syntax can improve precision and economy within expert communities. On the other hand, compression can increase reader burden by removing the explicit clausal scaffolding that helps non-specialists, interdisciplinary readers, and public audiences follow scientific claims. The substantive communication issue is therefore not whether LLM-polished prose is fluent, but whether fluency is being achieved through a form of compression that makes science less accessible outside expert circles.

Because the cross-register comparison and productivity analysis are presented in the Results, the implication here is interpretive: LLMs may be changing science in two opposite directions at once. They may make academic writing easier to produce for authors who face English-language writing demands, while also making the resulting prose more standardized, compressed, and distant from everyday language. This tension should caution against treating LLM-assisted writing as an unqualified improvement. Journals, reviewers, and authors may need to evaluate not only grammatical polish, but also communicative clarity, audience accessibility, and the norms of scientific expression that LLMs amplify.

### *Consequences for NNES scholars and participation in science*

For NNES scholars, LLMs may lower one of the most persistent barriers in English-language science. English constitutes the dominant medium of indexed scientific publication (Liu, 2017), while researchers writing outside their first language often face higher costs in reading, writing, presenting, and revising scientific work (Amano et al., 2023; Kourilova-Urbanczik, 2012; Lenharo, 2023). Our results suggest that LLM-assisted writing may help some NNES scholars produce prose that better matches the compressed norms of contemporary scientific English. In this sense, the linguistic shift we identify may have a participatory consequence: it could reduce some language-related costs of writing science in English.

This participatory potential should be interpreted cautiously. If LLMs help NNES scholars align with prevailing English-language norms, they may lower some barriers to publication while also reinforcing those norms as an implicit standard of scientific legitimacy. Moreover, access to high-quality LLM tools, prompt literacy, institutional policy, and journal attitudes toward AI assistance are themselves unevenly distributed. The same technology that reduces some writing costs may create new differences in who can use LLMs effectively, transparently, and acceptably.

### *Limitations and future directions*

Several limitations should guide interpretation. First, observational patterns cannot fully identify individual-level LLM usage, and current AI-text detection tools are insufficiently reliable for scientific writing. Second, author ethnicity and affiliation are imperfect proxies for language background, fluency, and writing practice, even though our stricter treatment-control design reduces some ambiguity. Third, our analyses focus on abstracts, which are highly compressed by genre and may not capture the full range of changes in introductions, methods, discussions, peer-review responses, or grant writing. Fourth, linguistic complexity is only one window into scientific language; future work should examine argument structure, hedging, evidentiality, novelty claims, and readability for different audiences.

Future research should move from detecting LLM traces to understanding how LLMs reshape the social and linguistic infrastructure of science. This requires direct evidence on author workflows, cross-linguistic comparisons beyond the groups examined here, analyses of full-text scientific articles, and experiments that evaluate not only linguistic complexity but also comprehension, trust, peer review, and public understanding. As LLMs become routine writing partners, the central question is no longer whether they are present in scientific writing, but how they are changing the language through which science is produced, judged, shared, and understood.

# Conclusion

This study shows that LLMs are reshaping the language of science, but its significance lies in what this linguistic shift may do to science itself. By accelerating compressed scientific English, LLMs may influence how knowledge is packaged, recognized as professional, and circulated through publication systems. Scientific language is not a neutral surface layer of research; it is part of the infrastructure through which claims become legible, credible, and authoritative.

The consequences are double-edged. LLM-assisted writing may broaden participation in scientific authorship by lowering some English-language writing barriers for NNES scholars and improving their ability to participate in global publication systems. At the same time, the same tools may narrow the accessibility of scientific communication by pushing prose toward greater lexical density, stronger terminological compression, and reduced explanatory scaffolding. In short, LLMs may help more people write science while making some scientific texts harder for broader audiences to read.

The question for the scientific community is therefore not only how to disclose or regulate LLM use, but what communicative norms these tools are helping to build. If LLMs are used primarily to make prose more publication-like, they may intensify the distance between expert discourse and broader understanding. If used deliberately, however, they could also support clearer and more inclusive scientific communication across linguistic backgrounds. The future of scientific writing will depend on whether LLMs become tools of compression alone, or tools for making scientific knowledge both more participatory and more understandable.

# Methodology

## *Dataset*

Our primary dataset derives from the November 1, 2024 snapshot of OpenAlex, a comprehensive scholarly database. We retrieved all English-language publications classified as 'article' (encompassing journal articles and conference proceedings), 'review', or 'preprint' published between January 1, 2020, and November 1, 2024. Following filtration for entries with at least one author and analyzable abstracts, we established a core corpus of 21.36 million publications. To contextualize long-term linguistic evolution, we supplemented this with abstracts from Nature, Science, and PNAS spanning 1950–2020 (375K articles).

Author ethnicity was classified using Ethnicseer, a name-based inference tool achieving 85% accuracy against bibliographic benchmarks (Treeratpituk & Giles, 2021). Institutional country affiliations were mapped via ROR (Research Organization Registry) identifiers, with ISO 3166-1 alpha-2 codes designating primary national associations. Publications were classified into domains (Life Sciences, Physical Sciences, Health Sciences, Social Sciences) and 26 subfields using OpenAlex's topic ontology, allowing multi-domain assignment per article. Journal Impact Factor (JIF) quartiles were computed per venue: mean citation counts (2020–2023 publications) were ranked within 252 OpenAlex subfields, with Q1 (top 25%) to Q4 (bottom 25%) assignments based on percentile thresholds. Comprehensive data processing protocols are documented in Appendix S2.

### *Metric Aggregation and Weighted Calculation Methodology*

Given the multidimensional nature of linguistic complexity metrics, we implement a two-tiered analytical approach to enhance interpretability. For primary trend visualization in Figure 1 panels (A, B, C, D, F, H), we aggregate granular indicators into composite lexical and syntactic complexity indices. Each text's lexical complexity percentile is computed as the mean percentile rank across all seven lexical metrics, while syntactic complexity percentile represents the average of fourteen syntactic metric percentiles. Formally, for text i:

$$Syntactic\ Complexity\ Percentile_i = \frac{1}{14}\sum_{j=1}^{14} P_{i,syn,j}$$

$$Lexical\ Complexity\ Percentile_i = \frac{1}{7}\sum_{j=1}^{7} P_{i,lex,j}$$

Here $P_{i,syn,j}$ denotes the percentile rank of text i for syntactic complexity indicator j (i.e., its relative position within the study sample). Analogous calculations apply to lexical complexity. This aggregation preserves multidimensional information while enabling cross-text comparability.

For country-level and ethnicity-level analyses (Figure 1F, 1H), we compute weighted complexity averages using author proportional representation. The national lexical complexity percentile for country c in year y is given by:

$$Country\ Lexical\ Complexity_{c,y} = \frac{\sum Author\ Share_{i,c,y} \times Lexical\ Complexity\ Percentile_{i,y}}{\sum Author\ Share_{i,c,y}}$$

with analogous calculation for syntactic complexity and ethnic linguistic complexity. Here, $Author\ Share_{i,j,y}$ represents the proportion of authors from country c in publication I (year y). The NES (Native English Speaking) category aggregates contributions from predominantly English-speaking nations: United States, United Kingdom, Canada, Australia, New Zealand, Ireland, Jamaica, and Trinidad and Tobago. This approach weights each publication's complexity by the proportional contribution of authors from ethnic j or country j. Due to international and cross-ethnic collaboration, trends may reflect influences from co-authors of other nationalities and ethnicities. Comprehensive computational details appear in Appendix S3.

### *Difference-in-Differences Methodology*

We employ a dual-level difference-in-differences (DID) framework to isolate the causal effect of large language models (LLMs) on linguistic complexity disparities between non-native (NNES) and native English-speaking (NES) scholars. Our analysis implements two complementary specifications to address distinct dimensions of the treatment effect. The first model examines journal-year level effects using ordinary least squares (OLS) estimation:

$$Y_{jit} = \alpha + \beta_1 CN_i \times PostGPT_t + \beta_2 CN_i + \eta X_i + Year_t + Journal_j + e_{jit}.$$

Where, $Y_{jit}$ represents the metrics of linguistic complexity for article $i$ published in journal $j$ during year $t$. The key treatment indicator $CN_i$ equals 1 for articles where all authors are Chinese-affiliated NNES scholars, and 0 for NES control groups. The post-period indicator $PostGPT_t$ takes the value 1 for publications in 2024 and 0 otherwise. The coefficient $\beta_1$ captures our primary parameter of interest, which is the differential change in linguistic complexity for NNES scholars following ChatGPT's introduction. The model controls for article-level covariates $X_i$ (citation counts, author numbers, open access status, and sentence counts) while including *Year* and *Journal* fixed effects to account for temporal trends and publication venue characteristics. $e_{jit}$ is a random error term.

The second specification analyzes effects at the author-year level:

$$Y_{ait} = \alpha + \beta_1 CN_i \times PostGPT_t + \beta_2 CN_i + \eta X_i + Year_t + Author_a + e_{ait}.$$

In the first specification,

The second model replaces Journal fixed effects with Author fixed effects to control for individual writing style persistence. Comprehensive design rationales, variable definitions, data processing protocols, and robustness checks are documented in Appendix S4.

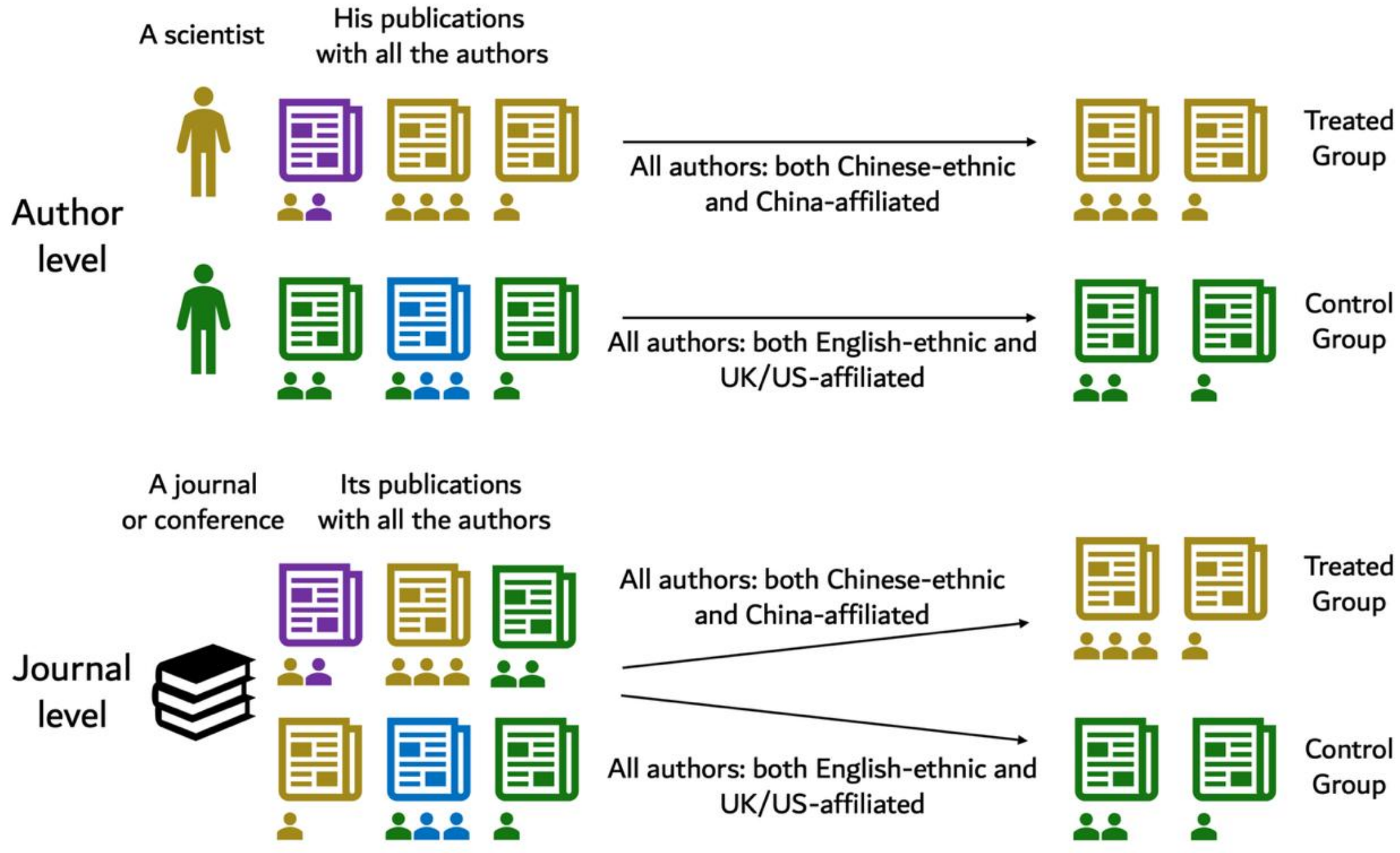

Fig 5. DID design quantifying post-ChatGPT divergence between treatment (NNES) and control (NES) groups relative to pre-adoption baselines. Treatment group: Publications where all authors are both ethnically Chinese and affiliated with Chinese institutions. Control group: Publications where all authors are both ethnically English and affiliated with US/UK institutions. Regression specifications incorporate first-author fixed effects and journal/conference fixed effects, enabling cross-venue and within-author comparisons of NNES vs. NES linguistic complexity shifts pre/post-ChatGPT adoption.

### *Experiment Design*

The experimental validation in Figure 3 utilized randomly sampled abstracts from our DID treatment group—articles exclusively authored by Chinese-ethnic researchers affiliated with Chinese institutions. We employed stratified random sampling without replacement to select three independent 1,000-abstract cohorts (Sample 1, Sample 2, Sample 3) from 2020 publications, ensuring temporal separation from widespread LLM adoption. All samples underwent controlled text-polishing procedures across four large language models (ChatGPT-4o, DeepSeek-V3, Qianwen-8B, LLaMA-4-Scout) with three instructional prompt variations, as detailed in Tables S7 (model specifications) and S8 (prompt formulations). Subpanel-specific protocols were implemented: Figure 3A processed Sample 1 under Prompt1 across all four LLMs; Figure 3B evaluated Samples 1-3 using ChatGPT-4o with Prompt1; Figure 3C tested Sample 1 under Prompt1 across four ChatGPT iterations (v3.5, v4, v4o, v-o1); and Figure 3D applied three distinct prompts (Prompt1-3) to Sample 1 via ChatGPT-4o. Linguistic complexity changes were quantified using paired t-tests comparing pre- versus post-polishing metrics across all 21 complexity dimensions, with extended methodological documentation provided in Appendix S5.

# Data, materials, and software availability

The data and code supporting this study are available at Zenodo: https://doi.org/10.5281/zenodo.21019659. Due to copyright and licensing restrictions, raw full-text data are not redistributed. Processed indicators, metadata, analysis scripts, and reproducible workflows are provided.

# Supplementary Information for

# Large language models reshape the language of science

This supplementary information (SI) file provides further detail and background information for "Large language models reshape the language of science." This SI considers each data source and each method used.
Numerous robustness tests to the main results are also provided.

# S1 Measurement framework

## S1.1 Indicator Selection and Conceptual Clarification

Numerous metrics exist for measuring linguistic complexity, alongside related but distinct concepts such as accuracy and fluency (Housen et al., 2012). This section clarifies our rationale for excluding these alternative metrics and emphasizes that we did not selectively report results favoring specific conclusions. Instead, after identifying all indicators meeting our research criteria, we faithfully report findings for every selected metric. Our selection criteria prioritized indicators that are: (1) highly interpretable, (2) robust to text length variation, and (3) empirically validated and widely adopted in subsequent research.

Within Second Language Acquisition (SLA) research, Fluency, Accuracy, and Complexity (FAC) are frequently employed as a triad to assess second language learners' proficiency (Housen et al., 2012). Accuracy quantifies deviations from native-like grammar and vocabulary usage, typically measured by counts of grammatical errors or misspellings (Lu et al., 2019; Nisioi, 2015). Fluency gauges the smoothness of language production, often assessed via metrics like speech rate (words per minute) or repair fluency (e.g., pause frequency/duration) (Chandler, 2003; Skehan, 2009; Tabari, 2016). While related, Accuracy and Fluency represent concepts parallel to, rather than synonymous with, linguistic Complexity. As our focus is specifically on Complexity, and Fluency metrics are particularly challenging to apply reliably to written scientific texts, these FAC dimensions were excluded from our framework.

We also excluded metrics based on Kolmogorov complexity and its derivatives. In linguistics, Kolmogorov complexity approximates the minimal description length of a text, often operationally measured using compression algorithms like gzip (Ehret, 2024; Juola, 1998; Nelson, 2024). This approach suffers from significant sensitivity to text length, rendering it unsuitable for our comparative analysis. Conceptually, Kolmogorov complexity aims to capture conciseness (expressing more meaning with fewer symbols). Aspects of this are partially addressed by our chosen indicators like lexical density and MTLD. Methodologically, reliance on gzip compression ratios yields results that lack direct linguistic interpretability.

Additionally, we did not adopt traditional readability formulas. Widely used metrics like the Dale-Chall formula (Dale & Chall, 1948; Nahatame, 2021) and others incorporate elements such as syllable counts, sentence length, proportion of complex words, or syntactic similarity. These formulas primarily measure readability– the ease with which text can be understood – rather than linguistic complexity. While related, the constructs differ. Some overlap exists; for instance, indicators we employ like word length and syllables per word are occasionally used in readability formulas (Sun et al., 2024). Generally, higher lexical and syntactic complexity correlates with lower readability.

## S1.2 Indicator Definitions and Computational Approaches

The indicator system is summarized in Supplementary Table S1.

**Lexical complexity** quantifies the sophistication of vocabulary usage in written texts. Following Lu's (Lu, 2012) framework, we measure three dimensions: lexical diversity, lexical sophistication, and lexical density.

**Lexical diversity** (also termed lexical variation) captures the range of distinct vocabulary employed. While Type-Token Ratio (TTR)—the proportion of unique words to total words—is widely used, it suffers from significant text-length bias (e.g., artificially inflated values in shorter texts) (McCarthy & Jarvis, 2010). Alternative corrections (e.g., Corrected TTR, Root TTR) partially mitigate this issue, but we adopt the Measure of Textual Lexical Diversity (MTLD) for its robustness. MTLD calculates the average span of text required to reach a predefined TTR threshold (0.72), providing length-invariant estimates of lexical richness (McCarthy & Jarvis, 2010). Its validation in applied linguistics and computational text analysis (Zenker & Kyle, 2021) makes it ideal for cross-linguistic and cross-author comparisons.

**Lexical sophistication** reflects the usage of advanced or low-frequency vocabulary (Read, 2009). One approach defines sophistication as the proportion of words outside a predetermined "common word list" (Laufer & Nation, 1995; Leech et al., 2001). However, this method lacks discriminative power in academic contexts due to discipline-specific terminology, and the selection of word list may cause bias. Instead, we use word length (mean characters per word) and syllables per word (mean syllables per word)—objective, interpretable metrics validated in SLA research (Ferris, 1994; Kormos, 2011; Lu et al., 2019) and widely adopted in NLP applications.

**Lexical density** measures the proportion of content words (nouns, verbs, adjectives, adverbs) to total words. Content words carry greater semantic weight than function words (e.g., prepositions, pronouns), making lexical density a proxy for information packaging efficiency (Ure, 1971). Higher density correlates with informational complexity, for example, academic writing exhibits greater density than spoken language (Révész & Brunfaut, 2013). We operationalize this via noun ratio, verb ratio, adjective ratio, and adverb ratio.

**Syntactic complexity** is assessed using Lu's (Lu, 2010; Lu, 2011) multidimensional framework, which integrates key constructs from SLA research and has been empirically validated (Housen et al., 2012; Jagaiah et al., 2020; Lu et al., 2019; Vajjala & Meurers, 2012). The framework comprises 14 indicators across five dimensions, computed using Kyle's (Kyle, 2016) syntactic parser. Core linguistic units are defined as follows:

- Words are total number of tokens that are not punctuation marks.
- A sentence (S) is a group of words delimited with one of the following punctuation marks that signal the end of a sentence: period, question mark, exclamation mark, quotation mark, or ellipsis.
- A clause (C) is defined as a structure with a subject and a finite verb, and includes independent clauses, adjective clauses, adverbial clauses, and nominal clauses.
- A dependent clause (DC) is defined as a finite adjective, adverbial, or nominal clause.
- A T-unit (T) is one main clause plus any subordinate clause or nonclausal structure that is attached to or embedded in it.
- A complex T-unit (CT) is one that contains a dependent clause.

- ➓ A coordinate phrase (CP) is a grammatical structure that connects related words, phrases, or clauses using coordinating conjunctions such as "and," "but," or "or." In Lu's definition, Only adjective, adverb, noun, and verb phrases are counted in coordinate phrases.
- ➓ Complex nominals (CN) comprise (i) nouns plus adjective, possessive, prepositional phrase, relative clause, participle, or appositive, (ii) nominal clauses, and (iii) gerunds and infinitives in subject position.
- ➓ Verb phrases (VP) comprise both finite and non-finite verb phrases.

The 14 indicators quantify five syntactic dimensions:

**Length of production units**: Mean length of clause (MLC), sentence (MLS), and T-unit (MLT)—positively correlated metrics reflecting structural elaboration.

**Sentence complexity**: Clauses per sentence (C/S), indexing clause embedding frequency.

**Subordination**: Clauses per T-unit (C/T), complex T-units per T-unit (CT/T), dependent clauses per T-unit (DC/T), and dependent clauses per clause (DC/C).

**Coordination**: Coordinate phrases per clause (CP/C), per T-unit (CP/T), and T-units per sentence (T/S).

**Particular structures**: Complex nominals per clause (CN/C), per T-unit (CN/T), and verb phrases per T-unit (VP/T).

Research comparing syntactic complexity between native speakers (NS) and non-native speakers (NNS) in L2 writing development reveals both consistent and divergent patterns across metrics. Most studies concur that higher proficiency learners and NS exhibit greater production unit length—evidenced by elevated scores in mean length of clause (MLC), sentence (MLS), and T-unit (MLT)—reflecting their tendency to employ longer syntactic structures with more embedded clauses (Ai & Lu, 2013; Cumming et al., 2005; Foster & Tavakoli, 2009; Ortega, 2003). Similarly, advanced NNS consistently demonstrate reduced usage of complex nominals relative to NS peers across investigations (Ai & Lu, 2013; Lu & Ai, 2015; nasseri., 2017). In contrast, findings for coordination (e.g., CP/C, CP/T, T/S), subordination (e.g., C/T, CT/T, DC/T), and sentence complexity (C/S) metrics display notable contradictions: some studies report NS superiority, others show NNS advantages, while additional work detects no significant group differences (Ai & Lu, 2013; Bardovi-Harlig, 1992; Lu & Ai, 2015; nasseri., 2017; Norris & Ortega, 2009). These inconsistencies likely stem from methodological variations across studies, including differences in learner L1 backgrounds, sample sizes, text genres (e.g., argumentative vs. descriptive writing), and proficiency thresholds used to classify "advanced" NNS.

# S2 Data Description

## S2.1 OpenAlex Data Source

Our primary dataset for scientific publications derives from OpenAlex, an open-access scholarly database developed by the nonprofit OurResearch. As one of the world's largest citation indexes,

OpenAlex comprehensively catalogs publications across journals, conference proceedings, books, datasets, and preprint servers. The database structures scholarly entities—including works, authors, sources, institutions, topics, publishers, and funders—alongside their interrelationships. For each publication, we extracted metadata comprising abstracts, publishing venues, author lists with institutional affiliations, publication dates, fields of study, citation counts, and open-access status. We utilized the November 1, 2024 snapshot of OpenAlex, restricting our analysis to publications from 2020 to 2024 inclusive.

To construct our analytical corpus, we implemented the following inclusion criteria: (i) publication year between 2020–2024; (ii) document type limited to 'article' (encompassing both journal and conference articles), 'preprint', and 'review'; (iii) English-language content; (iv) at least one listed author; and (v) presence of a valid abstract. To ensure abstract validity, we excluded entries with null values, removed boilerplate text (e.g., download prompts or duplicated placeholder content), and filtered abnormally short abstracts. Post-processing validation confirmed 99.6% functional accuracy of retained abstracts, with minor formatting artifacts (e.g., "Abstract:" prefixes) deemed acceptable. Our final baseline dataset comprises 21.36 million publications meeting these criteria.

To establish historical context for longitudinal comparison, we supplemented this corpus with abstracts from Nature, Science, and Proceedings of the National Academy of Sciences (PNAS) spanning 1950–2020. This supplementary dataset, extracted from OpenAlex, encompasses more than 375 thousand articles.

**Domain & Field Classification**. OpenAlex employs a four-tier hierarchical taxonomy: Domain → Field → Subfield → Topic. Each publication is tagged with relevant topics, enabling classification across these levels. We mapped publications to their primary Domain (Life Sciences, Physical Sciences, Health Sciences, or Social Sciences) and Field (26 distinct fields nested within the four Domains) based on topic associations. A single publication may belong to multiple Domains or Fields. The full classification schema and field-domain mappings are detailed in Table S2 and the OpenAlex documentation.

**Journal Impact Factor (JIF) Quartiles**. Given that standard Journal Citation Reports (JCR) quartiles cover only ~10,000 Web of Science-indexed journals—while our dataset spans over 90,000 venues—we computed custom quartile rankings using OpenAlex data. We calculated JIF as the mean citation count for articles published in a venue (journal or conference) between 2020–2023. Venues were then ranked within their respective OpenAlex subfields (252 total), with quartiles assigned as follows: Q1 (top 25%), Q2 (25–50%), Q3 (50–75%), and Q4 (75–100%). For venues associated with multiple subfields, we assigned a unified quartile by: (1) selecting the subfield with the most frequent topic associations as primary, or (2) if topic frequency was equal across subfields, retaining the highest quartile. Venues lacking publications during 2020–2023 or missing topic labels were excluded from quartile assignment.

**Institution and Country Mapping**. Institutional affiliation data were sourced directly from OpenAlex, which utilizes the Research Organization Registry (ROR; https://ror.org) for unique identifiers. Each institution was mapped to its primary country of operation using ISO 3166-1 alpha-2 codes. Country code-name correspondences for all nations referenced in this study are provided in Table S3.

## S2.2 Ethnicity Detection via Ethnicseer

Ethnicseer was prioritized for name-based author-group classification due to its validated efficacy in bibliometric analysis. This tool employs a Wikipedia-derived name-to-group inference model (Treeratpituk & Giles, 2021), achieving 85% accuracy overall with 99% precision and 89% recall in computer science bibliographies—performance metrics critical for handling OpenAlex's multidisciplinary scope. Ethnicseer categorizes names into 12 name-inferred groups, with abbreviations and full group labels detailed in Table S4. We applied this tool to author names in OpenAlex to determine name-inferred group probabilities.

# S3 Statistical Methodology

## S3.1 Calculation of Mean Linguistic Complexity Percentiles

Given the multidimensional nature of linguistic complexity metrics, we aggregated specific indicators to the lexical complexity and syntactic complexity levels for analytical clarity. For each text, we computed percentile ranks for all lexical complexity indicators (or syntactic complexity indicators) relative to the full sample. The mean percentile for each complexity dimension was then derived by averaging these individual indicator percentiles:

$$Syntactic\ Complexity\ Percentile_i = \frac{1}{m}\sum_{j=1}^{m} P_{i,syn,j}$$

$$Lexical\ Complexity\ Percentile_i = \frac{1}{n}\sum_{j=1}^{n} P_{i,lex,j}$$

Here $P_{i,syn,j}$ denotes the percentile rank of text i for syntactic complexity indicator j (i.e., its relative position within the study sample), with m=14 representing the total syntactic complexity metrics. Analogous calculations apply to lexical complexity (n indicators). This aggregation method underpins results in Figure 1A, 1B, 1C, 1D, 1F, and 1H.

## S3.2 Linguistic Complexity Evolution: 1950–2020

We analyzed more than 375,000 article abstracts from Nature, Science, and PNAS spanning 1950–2020. Percentile ranks for all complexity metrics were computed exclusively within this historical corpus. Figure 1A presents aggregated trends, while Appendix Figures S1 and S2 detail temporal trajectories for individual lexical and syntactic complexity indicators, respectively.

## S3.3 Linguistic Complexity Evolution: 2020–2024

For the primary corpus of 21.36 million abstracts (2020–2024), we computed percentile ranks within this full sample before aggregating to lexical/syntactic complexity levels per Section 3.1. Figure 1B displays annual trends in aggregated complexity, with Appendix Figure S3 showing indicator-specific dynamics. Unless otherwise specified, all subsequent percentile calculations reference rankings within the 21.36-million-abstract sample.

## S3.4 Domain-Specific Trends

Publications were classified by domain/field using OpenAlex topic mappings (Section S2.1). After excluding articles with missing topic data (final n=21.2million), Figure 1C illustrates domain-stratified trends, while Appendix Figure S4 provides field-level granularity.

## S3.5 Journal Impact Factor (JIF) Quartile Trends

Venues were assigned JIF quartiles per Section S2.1. Analyses excluded publications from venues without quartile assignments (final n=19.23million). Figure 1D visualizes complexity trends stratified by venue quartile.

## S3.6 National Contribution Shares

Country-level publication shares (Figure 1E) were computed as:

$$Country\ Publication\ Share_{c,y} = \frac{\sum Author\ Share_{i,c,y}}{Global\ Total\ Publications_{y}}$$

where: $Author\ Share_{i,c,y}$ is proportion of authors from country c in publication i published in year y, $Global\ Total\ Publications_{y}$ is total publications number in year y. After excluding publications with missing affiliation data (final n=17.11million), contributions were summed across all publications. Since an author may belong to multiple institutions and different countries, this author is included in the calculations for each respective country's author share. The NES (Native English Speaking) category aggregates contributions from predominantly English-speaking nations: United States, United Kingdom, Canada, Australia, New Zealand, Ireland, Jamaica, and Trinidad and Tobago.

## S3.7 Country-Level Linguistic Complexity

National complexity metrics (Figure 1F) were calculated as author-share-weighted averages:

$$Country\ Lexical\ Complexity\ Percentile_{c,y} = \frac{\sum Author\ Share_{i,c,y} \times Lexical\ Complexity\ Percentile_{i,y}}{\sum Author\ Share_{i,c,y}}$$

$$Country\ Syntactic\ Complexity\ Percentile_{c,y} = \frac{\sum Author\ Share_{i,c,y} \times Syntactic\ Complexity\ Percentile_{i,y}}{\sum Author\ Share_{i,c,y}}$$

This approach weights each publication's complexity by the proportional contribution of authors from country c. Due to international collaboration, trends may reflect influences from co-authors of other nationalities.

### S3.8 Name-Inferred Author-Group Contribution Shares

Name-inferred author-group publication shares (Figure 1G) follow an analogous formula:

$$Ethinic\ Publication\ Share_{e,y} = \frac{\sum Author\ Share_{i,e,y}}{Global\ Total\ Publications_y}$$

where $Author\ Share_{i,e,y}$ represents the proportion of authors in name-inferred group e in publication i published in year y. Name-inferred group classifications (e.g., ENG = English-heritage) correspond to Table S4.

### S3.9 Name-Inferred Author-Group Linguistic Complexity

Name-inferred author-group-stratified complexity (Figure 1H) was computed as author-group-share-weighted averages:

$$Ethnic\ Lexical\ Complexity\ Percentile_{e.y} = \frac{\sum Author\ Share_{i,e,y} \times Lexical\ Complexity\ Percentile_{i,y}}{\sum Author\ Share_{i,e,y}}$$

$$Ethnic\ Syntactic\ Complexity\ Percentile_{e,y} = \frac{\sum Author\ Share_{i,e,y} \times Syntactic\ Complexity\ Percentile_{i,y}}{\sum Author\ Share_{i,e,y}}$$

This approach weights each publication's complexity by the proportional contribution of authors from ethnic e. As with country-level analyses, these weighted averages may be influenced by co-authors of other name-inferred author groups in collaborative works.

# S4 Methodological Details for Difference-in-Differences (DID) Analysis

## S4.1 Treatment and Control Group Construction

Accurately distinguishing native English-speaking (NES) and non-native English-speaking (NNES) author groups presents challenges when relying solely on either name-based group inference or institutional affiliations. Potential misclassifications include: international researchers employed at Chinese institutions, French scholars at U.S. universities, authors using Anglicized names, or legally changed surnames. To mitigate these issues, we implemented a dual-verification protocol:

Treatment Group (NNES): Articles where all authors are both classified as Chinese by name inference and affiliated with Chinese institutions.

Control Group (NES): Articles where all authors are both classified as English by name inference and affiliated with U.S./U.K. institutions.

This approach minimizes contamination from mixed author-group collaborations. For robustness checks, we expanded the treatment group to include seven additional homogeneous cohorts:

Japanese: All authors ethnically Japanese + Japan-affiliated

Indian: All authors ethnically Indian + India-affiliated

German: All authors ethnically German + Germany-affiliated

French: All authors ethnically French + France-affiliated

Italian: All authors ethnically Italian + Italy-affiliated

Korean: All authors ethnically Korean + Korea-affiliated

Russian: All authors ethnically Russian + Russia-affiliated

The primary NNES vs. NES comparison (2.47M NNES vs. 331K NES articles) underpins Figure 2D, while multi-group results appear in Figure S6. Final sample size after name-based and institutional filtering: 2,801,218 articles.

## S4.2 Variable construction

All variables are defined at the article level:

CN (Treatment Indicator): Binary variable where 1 = article authored exclusively by name-inferred Chinese researchers affiliated with Chinese institutions (NNES), 0 = article authored exclusively by name-inferred English researchers affiliated with U.S./U.K. institutions (NES).

**PostGPT (Time Indicator)**: Binary variable marking ChatGPT adoption period (1 = 2024 publications; 0 = 2020–2023 publications). The 2024 cutoff accounts for publication lags following ChatGPT's November 2022 release.

**Citation Counts (Control variables)**: Total citations received by each article as of November 1, 2024, controlling for scholarly impact.

**Number of Authors (Control variables)**: Count of listed authors, adjusting for collaboration effects.

**Is_OA (Control variables)**: Binary indicator indicates Open Access Status (1 = open access; 0 = paywalled).

**Number of Sentences (Control variables)**: Sentence count per abstract.

**JIF Quartile**: Venue-level quartile ranking (Q1–Q4) computed per methodology in Appendix S2.1.

**Year Fixed Effects**: Dummy variables for each publication year (2020–2024).

**Journal Fixed Effects**: Unique identifiers for each publication venue.

**Author Fixed Effects**: First author identifiers capturing individual writing styles.

## S4.3 Econometric Models

We employ a difference-in-differences (DID) framework to estimate the causal effect of ChatGPT's emergence on linguistic complexity in non-native English speaker (NNES) academic writing. The analysis uses two complementary model specifications to capture different dimensions of the treatment effect.

***Journal-Year Level Analysis***. The first model examines effects at the journal-year level using ordinary least squares (OLS) estimation:

$$Y_{jit} = \alpha + \beta_1 CN_i \times PostGPT_t + \beta_2 CN_i + \eta X_i + Year_t + Journal_j + e_{jit}.$$

In this specification, $Y_{jit}$ represents the metrics of linguistic complexity for article $i$ published in journal $j$ during year $t$. The key treatment indicator $CN_i$ equals 1 for articles where all authors are Chinese-affiliated NNES scholars, and 0 for NES control groups. The post-period indicator $PostGPT_t$ takes the value 1 for publications in 2024 (the ChatGPT adoption period) and 0 otherwise. The coefficient $\beta_1$ captures our primary parameter of interest, which is the differential change in linguistic complexity for NNES scholars following ChatGPT's introduction. The model controls for article-level covariates $X_i$ (citation counts, author numbers, open access status, and sentence counts) while including *Year* and *Journal* fixed effects to account for temporal trends and publication venue characteristics. The treatment units consist of all NNES articles published in journal $j$ during year $t$, compared against NES articles in the same journal-year contexts.

***Author-Year Level Analysis.*** The second specification analyzes effects at the author-year level:

$$Y_{ait} = \alpha + \beta_1 CN_i \times PostGPT_t + \beta_2 CN_i + \eta X_i + Year_t + Author_a + e_{ait}.$$

Here, $Y_{ait}$ measures metrics of linguistic complexity for article i written by author $a$ in year t. This model replaces Journal fixed effects with Author fixed effects to control for individual writing style persistence. The treatment group comprises all articles published by NNES author $a$ in year $t$, compared against publications by NES authors during the same years. This specification helps isolate ChatGPT's effect on individual writing patterns while accounting for author-level heterogeneity.

## S4.4 Data Description for DID Analysis

Descriptive statistics for the DID treatment (Chinese NNES) and control (English NES) groups are presented in Tables S5 (treatment group) and S6 (control group). Figure S5 shows the annual percentile means of specific indicators for the treatment group and the control group. The Chinese sample size significantly exceeds the NES group (2.47M vs. 331K articles), reflecting China's homogeneous ethnic composition versus the ethnically diverse U.S. scholarly landscape. Consequently, mean team size differs substantially: 5.53 authors per publication in the Chinese sample versus 1.74 in the U.S. sample. This discrepancy partly stems from disciplinary variations—social sciences and humanities (predominant in the U.S. sample) typically feature smaller teams than STEM fields (which were disproportionately filtered out by our homogeneous ethnicity requirement).

Missing values for disciplinary classification (Domain/Field) occur when publication venues lack topic labels in OpenAlex. Similarly, JIF quartile assignments are missing when venues lack subfield classifications. Computational details are documented in Appendix S2.1.

All regression analyses use raw metric values rather than percentiles. To address scale disparities, three length-based syntactic complexity metrics—Mean Length of Clause (MLC), Sentence (MLS), and T-unit (MLT)—were divided by 10, 20, and 10 respectively before regression. Coefficient estimates for these rescaled variables should be interpreted with corresponding 10×, 20×, and 10× multipliers; this transformation does not affect statistical significance or coefficient direction.

## S4.5 Robustness Checks and Heterogeneity Analysis

We conducted split-sample regressions and alternative specifications to assess robustness and heterogeneity.

**Alternative Treatment Timing (After_2023).** While ChatGPT-3.5 launched in November 2022, widespread academic adoption likely began in 2023. Figure 1 shows early complexity shifts emerging in 2023, though less pronounced than 2024's discontinuity. We therefore defined an alternative temporal indicator: *After_2023*= 1 for publications ≥2023 (0 otherwise). Results (Appendix Fig S6) show non-significance for most metrics despite including 2024 data, confirming that ChatGPT's linguistic impact became statistically detectable only in 2024. This lag likely reflects adoption timelines, peer-review delays, and publication cycles rather than version-specific linguistic differences—a finding corroborated by our experiments in Figure 3.

**Constituent Syntax Unit Analysis.** To address concerns that syntactic complexity metrics might obscure granular changes in specific grammatical structures, we regressed the DID model against constituent units (e.g., clause counts, coordinate phrases). Results (Fig S7) show no significant changes in any underlying syntactic unit (all $p \geq 0.05$), confirming that ChatGPT does not systematically alter specific grammatical elements (e.g., increasing subordinate clauses or reducing coordination) in total. Abstract length (measured by sentence/word counts) also remained stable.

**Percentile Rank Regression.** Our primary analyses use raw metric values for interpretability. For alignment with Figure 1's percentile-based visualizations, we replicated the DID analysis using percentile ranks as dependent variables. Results (Fig S8) show consistent effect directions and significance levels with our main findings (Fig 2) and experimental results (Fig 3).

Author-Group-Specific Effects. Applying identical screening criteria (Section 4.1) to other author groups reveals differential impacts (Fig S9). Japanese and Korean author groups, whose dominant language backgrounds differ more typologically from English, show effect magnitudes similar to those observed in the Chinese NNES sample. German, French, and Russian author groups show minimal or inconsistent divergence from NES controls, while Italian and Indian author groups display intermediate effects.

**Stratification by Journal Impact Factor (JIF) Quartile.** Quartile-stratified regressions (Fig S10) used journal-level specifications due to limited author-level observations for each author in subsamples. All quartiles show statistically significant effects, though Q1 journals demonstrate the strongest significance (possibly due to larger sample sizes). While coefficient magnitudes vary, direct cross-quartile comparisons are statistically inappropriate. Combined with Figure 1D trends, we conclude ChatGPT's impact permeates all journal tiers, indicating universal adoption across academia's prestige spectrum.

**Discipline-Specific Analysis.** Domain-stratified regressions (Fig S11) reveal field-level nuances. Though effect sizes differ across Life Sciences, Physical Sciences, Health Sciences, and Social

Sciences (field-domain mappings in Table S2), all domains exhibit the core pattern: lexical complexity increases and syntactic complexity decreases. These findings align with domain-level trends in Figure 1C.

# S5 Experimental Validation Methodology

## S5.1 Chinese NNES Sample Experiment

The experimental validation in Figure 3 utilized randomly sampled abstracts from our DID treatment group—articles exclusively authored by name-inferred Chinese researchers affiliated with Chinese institutions. We drew three independent samples of 1,000 abstracts from 2020 publications by random selection without replacement. All samples were processed through LLM polishing pipelines under controlled conditions. Full model details in Table S7. Prompt engineering variations documented in Table S8.

Subpanel-specific protocols:

Fig 3A: Sample 1 + Prompt1 + four LLMs (ChatGPT-4o, DeepSeek-V3, Qianwen-8B, LLaMA-4-Scout);

Fig 3B: Samples 1/2/3 + Prompt1 + ChatGPT-4o;

Fig 3C: Sample 1 + Prompt1 + four ChatGPT versions (v3.5, v4, v4o, v-o1);

Fig 3D: Sample 1 + three prompts (Prompt1/2/3) + ChatGPT-4o.

Linguistic complexity changes were quantified using paired t-tests comparing pre- vs. post-polishing metrics.

## S5.2 U.S./U.K. NES Sample Experiment

A parallel methodology was applied to the DID control group—articles authored exclusively by English-ethnic researchers affiliated with U.S./U.K. institutions. Three independent samples (1,000 abstracts each) were randomly selected without replacement from 2020 publications. Identical polishing procedures (models, prompts, statistical tests) from Section 5.1 were implemented. Results are presented in Figure S12.

## S5.3 Cross-Group Comparison

Fig S12A examines inherent linguistic differences between Chinese NNES and U.S./U.K. NES samples via independent t-tests comparing unpolished abstracts. Results show strikingly consistent complexity gaps across all three sample pairs ($p<0.001$), confirming robust sample-specific writing patterns.

Fig S12B assesses ChatGPT-4o polishing effects (Prompt1) on both cohorts. While subtle variations in coefficient magnitudes exist, the core finding—pre-existing linguistic differences between Chinese NNES and U.S./U.K. NES sample texts far outweigh polishing-induced changes—holds universally across samples. This demonstrates that LLM polishing cannot bridge pre-existing differences in writing patterns associated with language background and publication context.

# S6 Supplemental Analysis

## S6.1 Word Length Dynamics

Given the widespread use and interpretational clarity of word length as a linguistic metric—along with its exceptional statistical robustness in our findings—we conduct a dedicated difference-in-differences (DID) analysis of this indicator.

**Data and Methodology.** This supplementary analysis utilizes the same dataset as our primary DID regressions (Appendix S4), with word length isolated as the focal dependent variable. Data processing, variable definitions, and model specifications follow identical procedures documented in S4. To enhance analytical rigor, we implement an event study design using the following specifications:

$$Word\ Length_{jit} = \alpha + \sum_t \beta_t CN_i \times Year_t + \beta_2 CN_i + \eta X_i + Year_t + Journal_j + e_{jit}.$$

$$Word\ Length_{ait} = \alpha + \sum_t \beta_t CN_i \times Year_t + \beta_2 CN_i + \eta X_i + Year_t + Author_a + e_{ait}.$$

These models replace the singular post-period indicator with year-specific interaction terms (t= 2020, 2021, 2022, 2024), establishing 2023 as the pre-ChatGPT reference year. The coefficients $\beta_t$ capture dynamic treatment effects over time. $Word\ Length_{jit}$ denotes the word length metric for abstract I published in venue j during year t. All other variables retain their definitions from S4.3.

**Data Visualization.** Figure S14 illustrates the temporal evolution of lexical complexity, measured by Word Length percentile ranks across groups. Prior to 2024, both NNES and NES articles exhibited parallel trajectories in lexical complexity. This pattern diverged sharply in 2024, coinciding with the widespread adoption of ChatGPT in academic writing workflows. NNES abstracts surged to a 7-percentile lead over controls, reflecting a 6-percentile increase from 2023 levels, while NES abstracts showed consistent trends. The abrupt acceleration in NNES lexical complexity aligns temporally with ChatGPT's integration into research practices, contrasting with the stability observed in NES cohorts. These patterns suggest a differential response to language model tools between NNES and NES.

**Primary Results.** Table S9 demonstrates a statistically significant increase in word length following ChatGPT's introduction to NNES academic writing. Across all specifications, the CN×PostGPT interaction shows positive coefficients (0.080–0.114; $p<0.001$), persisting under increasingly stringent controls. Peak effects emerged in author fixed effects models ($\beta=0.114$, $p<0.001$), indicating NNES scholars systematically adopted longer words irrespective of individual writing tendencies. Journal fixed effects models ($\beta=0.094$, $p<0.001$) confirm this complexity increase occurred within the same journals, eliminating venue-specific biases. The combined model maintains strong significance ($\beta=0.097$, $p<0.001$, $R^2=0.707$), demonstrating ChatGPT's impact transcends both journal norms and author characteristics. Practical significance is notable: a 0.1 coefficient increase corresponds to approximately half an additional letter per word in NNES writing.

**Event Study Analysis.** Figure S15 presents dynamic treatment effects from journal-level (Panel a) and author-level (Panel b) event studies. Results validate parallel pre-trends, with 2020–2022 coefficients hovering near zero. The 2024 effect surge contrasts sharply with this stability, confirming ChatGPT's causal impact. Negative coefficients for 2021–2022 (relative to 2023) reflect an artifact of early adoption: manuscripts drafted with ChatGPT shortly after its November 2022 release could undergo accelerated editorial workflows, appearing in 2023 publications. Consequently, 2023 outputs represent a mixture of pre- and post-LLM practices, whereas 2024 publications more fully reflect ChatGPT-integrated workflows.

**Heterogeneity Analysis.** Journal Tier Stratification (Table S10) reveals nuanced variation, with marginally stronger effects in Q2 journals ($\beta$=0.154 vs. Q1$\beta$=0.085). Differences between tiers remain modest, however, indicating ChatGPT's complexity-enhancing effects transcend journal quality categories. Caution: Direct coefficient comparisons are statistically inappropriate due to varying sample sizes across quartiles and split-sample design. Domain Stratification (Table S11) shows disciplinary variation, with strongest effects in social sciences ($\beta$=0.150, $p<0.001$) and life sciences ($\beta$=0.134, $p<0.01$), followed by physical sciences ($\beta$=0.110, $p<0.001$) and health sciences ($\beta$=0.074, $p<0.01$). Notwithstanding these differences, effects persist across all domains, reinforcing ChatGPT's broad academic impact. Caution: Cross-coefficient comparisons are limited by differential sample sizes and split-sample design.

**Synthesis.** Word length analyses corroborate our core findings: ChatGPT-induced linguistic transformations initiated in 2023 but manifested prominently in 2024. Table S9 demonstrates the robustness of word lengthening effects, confirming LLMs amplify lexical complexity through extended vocabulary. Heterogeneity analyses further validate result stability while revealing subtle contextual variations. Collectively, this supplemental investigation reinforces our principal conclusion: ChatGPT's transformation of scientific writing is pervasive, significantly impacting every discipline and journal tiers.

## S6.2 Characteristic Regression Analysis

Data and Methodology. To investigate nuanced linguistic patterns across author groups, we conducted OLS regressions where the proportion of authors from specific author groups served as independent variables, and linguistic complexity metrics served as dependent variables. After selecting papers from the full corpus published between 2020 and 2023, we obtained a dataset encompassing 15.64 million publications excluding LLM-affected publications in 2024. and then estimated the following model:

$$Y_i = \alpha + \beta_1 Ethnic_Ratio_i + \eta X_i + Year_i + e_i$$

Where $Y_i$ is the linguistic complexity metric for publication i. $Ethnic_Ratio_i$ is proportion of authors belonging to specific name-inferred author-group and country-affiliation pairs:

- name-inferred Chinese authors at Chinese institutions (CN/CHI)
- name-inferred English authors at U.S./U.K. institutions (US/ENG)
- name-inferred Japanese authors at Japanese institutions (JP/JAP)
- name-inferred German authors at German institutions (DE/GER)
- name-inferred Indian authors at Indian institutions (IN/IND)

- name-inferred French authors at French institutions (FR/FRN)

$X_i$ is a group of control variables including citation counts, author numbers, open access status, and sentence counts. $Year_i$ is the year fixed effects.

Key findings. Fig S16 reveals distinct linguistic signatures across author groups relative to global baselines. For lexical complexity, Chinese, English, and German author groups exhibit consistently higher lexical sophistication, potentially reflecting preferences for technical terminology. Japanese and Indian author groups demonstrate comparatively lower lexical complexity. For syntactic complexity, English NES sample authors produce significantly more complex syntactic structures, while Japanese, German, Indian, and French author groups show marked simplification. This figure depicts a very nuanced pattern of the linguistic complexity differences across author groups.

## S6.3 Accuracy Assessment of AI Detection Tools for Scientific Text

We evaluated four AI-generated text detection tools to determine their efficacy in identifying large language model (LLM)-polished academic manuscripts. Our test corpus comprised:

- Human-written samples: 100 original abstracts from 2020 publications randomly selected from our DID dataset.
- AI-polished samples: 400 polished versions generated by processing each human abstract through four LLMs (ChatGPT-4o, DeepSeek, LLaMA, Qianwen).

Detection was performed using Python's PyDetectGPT package implementing four methods: DetectLLM (Su et al., 2023), FastDetectGPT (Bao et al., 2024), Log-Likelihood, and Log-Rank. Results are presented in Table S12.

Critical findings reveal fundamental limitations.

High false positive rates: Detection tools incorrectly flagged 2-18% (12.75% on average) of human-written abstracts as AI-generated.

Critical false negatives: For AI-polished texts, tools misclassified 72-100% (85.56% on average) as human-written across all LLMs.

Methodological failure: Detection rates showed negligible differences between human and AI-polished texts.

These results demonstrate that current detection tools lack statistical power to distinguish LLM-polished scientific writing from human-authored content.

## S6.4 Comparison of Academic and Everyday Texts

To benchmark the linguistic profile of scientific abstracts, we compared them against texts from two non-specialist registers: news and informal online discourse. News text was represented by abstract data from The New York Times, a register positioned between conversational and academic language. The data was sourced from the "The New York Times Articles Metadata" dataset on Kaggle, which contains a comprehensive collection of over 2.1 million article metadata entries from January 1, 2000, onward (*The New York Times Articles Metadata*, 2025). Our analysis utilized all available articles from 2000 to 2024.

For everyday conversational text, we employed the Reddit Corpus from Cornell University's Convokit (Chang et al., 2020). This dataset provides all queries and responses from all subreddits between 2008 and 2018. We specifically extracted dialogues from the r/AskReddit subreddit. Due to the platform's exponential data growth, we sampled up to one million entries per year; for years with fewer than one million total entries, all available data was included.

Both the New York Times summaries and the Reddit texts underwent the same cleaning and filtering procedures applied to our core dataset. The criteria were: (i) retention of English-language content only; and (ii) presence of a valid abstract. To ensure abstract validity, entries with null values were excluded, boilerplate text (e.g., download prompts or duplicated placeholder content) was removed, and abnormally short abstracts were filtered out.

The scientific text corpora consisted of: 1) all abstracts from 2000–2024 for the journals Nature, PNAS, and Science; and 2) all eligible abstracts from 2020–2024 from our main OpenAlex corpus, excluding papers from the aforementioned three journals (i.e., the primary dataset used in the main analysis).

We calculated lexical and syntactic complexity scores for every text across all four corpora. Each text was then ranked within the entire pooled dataset for each complexity dimension. The average percentile rank of a text across both complexity measures was computed to represent its overall relative linguistic complexity. Finally, texts were grouped by their register (scientific, news, online forum) and year, and the mean relative rank for each group was calculated to establish their aggregate positions on the complexity continuum.

The results, illustrated in Figure 4A, reveal a distinct hierarchy. Scientific papers exhibit significantly higher lexical complexity than news and internet texts, with a notable surge observed in 2024. Abstracts from top journals like Nature, Science, and PNAS show slightly higher average lexical complexity than other scientific papers. New York Times news texts are lexically more complex than internet texts, with the latter registering the lowest complexity. Over time, the lexical complexity of scientific texts has gradually increased, widening its gap with non-specialist registers. Conversely, as seen in Figure 4C, while the syntactic complexity of scientific texts remains higher than that of news and internet texts, it demonstrates a declining trend. The syntactic complexity of internet and news texts is comparable and relatively low. Notably, abstracts from top journals exhibit lower syntactic complexity than other scientific paper abstracts. This pattern suggests that the reduction in syntactic complexity from an initially high baseline may correspond to a compression of clausal structures into denser phrasal units.

### S6.5 Empirical Analysis of Associations Between Linguistic Complexity Changes and NNES-NES Publication Output Differences

To examine whether the use of large language models (LLMs) leads to an increase in publication productivity for NNES authors, we construct a regression model with an interaction term. The regression specification is as follows:

$$\Delta Production_a = \beta_0 + \beta_1(\Delta Complexity_a * NNES_a) + \beta_2 \Delta Complexity_a + \beta_3 NNES_a + \gamma X_a + \epsilon_a$$

The dependent variable is defined as the difference between each author's number of publications in 2024 and their average annual number of publications during 2020-2023, which captures the change in productivity before and after the emergence of LLMs. To quantify the use of LLMs, we employ the change in language complexity of publications, denoted as $\Delta Complexity_a$ in the regression equation. Specifically, this variable is calculated as the difference between the average language complexity of all papers published by an author in 2024 and the average language complexity of all papers published by the same author during 2020-2023. In the empirical analysis, we alternatively use changes in lexical complexity and syntactic complexity as the key explanatory variable. $NNES_a$ is a binary variable equal to 1 for non-native English-speaking authors (NNES, Chinese authors in this sample) and 0 for native English-speaking authors (NES). Based on this, we construct an interaction term between $NNES_a$ and $\Delta Complexity_a$ to capture the differential effect of LLM-related language complexity changes on publication productivity between NNES and NES authors, which is the main variable of interest in this study. $X_a$ represents a set of control variables, including academic age, the number of publications before 2019 (as including post-2020 publications would introduce severe multicollinearity), the number of citations before 2019, and the share of each author's publications in four broad fields (Life Science, Health Science, Physical Science, and Social Science), which are included to control for differences in disciplinary composition that may affect publication productivity.

Table S13 presents the baseline regression results. Columns (1) and (2) report the regression results using lexical complexity and syntactic complexity as the key explanatory variable, respectively. The results show that, in terms of publication output, NNES authors have lower baseline output than NES authors in this specification, as indicated by the negative coefficient on the NNES variable. However, an increase in lexical complexity is associated with higher annual publication output for NNES authors relative to NES authors by 0.0013 ($p < 0.001$), while the increase in language complexity itself is not associated with an increase in publication output (the coefficient on $\Delta Complexity_a$ is negative). In contrast, when syntactic complexity is used (Column 2), its increase is not associated with an improvement in publication output for NNES authors relative to NES authors, as the coefficient is not statistically significant.

To provide a more intuitive illustration of the results, we divide authors into three groups according to the terciles of changes in language complexity before and after 2024, and conduct grouped regressions. The regression specification is as follows:

$$\Delta Production_a = \beta_0 + \beta_1 NNES_a + \beta_2 Group2_a + \beta_3 Group3_a + \beta_4(NNES_a * Group2_a) + \beta_5(NNES_a * Group3_a) + \gamma X_a + \epsilon_a$$

The regression results are presented in Figure 4B, 4D. As shown in Figure 4, as the lexical complexity of NNES authors' papers increases, the gap in publication output between NNES and NES authors correspondingly narrows, which is consistent with our baseline regression results. In contrast, syntactic complexity exhibits a pattern of first decreasing and then increasing, which may help explain why the regression results for syntactic complexity were not statistically significant in the previous analysis.

We then conduct robustness check and heterogeneity analyses. First, we replace the dependent variable with the change in each author's publication rank, defined as the difference between the author's publication rank in 2024 and their average publication rank during 2020-2023. The regression results are reported in Table S14. The results are consistent with the baseline regression findings: relative to NES authors, an increase lexical complexity for NNES authors leads to an improvement in their publication rank (as indicated by a negative and statistically significant coefficient).

After modifying the regression specification to the same grouped regression model, the results remain consistent. As language complexity increases, the relative ranking difference between NNES and NES authors gradually declines. In addition, as lexical complexity increases, this relative difference is no longer statistically significant.

Figure S18 reports the regression results by author group. The estimates for German authors are similar to those for the Chinese NNES sample, with slightly larger coefficients. For Japanese authors, the increase in publication output relative to NES authors associated with language complexity is mainly driven by changes in syntactic complexity. For French authors, whose dominant language background is more typologically similar to English, none of the three regressions are statistically significant, suggesting that the association between LLM-related linguistic change and relative publication output is not statistically significant for this group.

We then sequentially replace the dependent variable with the change in the number of publications in journals of different quartiles to examine the heterogeneous effects of changes in language complexity across journal tiers. The results are shown in Figure S19. The figure shows clear heterogeneity between publications in Q1 journals and those in Q2, Q3, and Q4 journals. An increase in lexical complexity is associated with an increase in the number of publications by NNES authors in Q1 journals (relative to NES authors). In contrast, syntactic complexity shows the opposite association with NNES authors' publication output, corresponding to a decrease in publications in Q1 journals and an increase in publications in Q2-Q4 journals.

Figure S20 presents heterogeneity across fields. The results show that only in Health Sciences is an increase in lexical complexity associated with higher publication output for NNES authors relative to NES authors, while the results in the other fields are not statistically significant.

# S7 Supplementary Information Figures

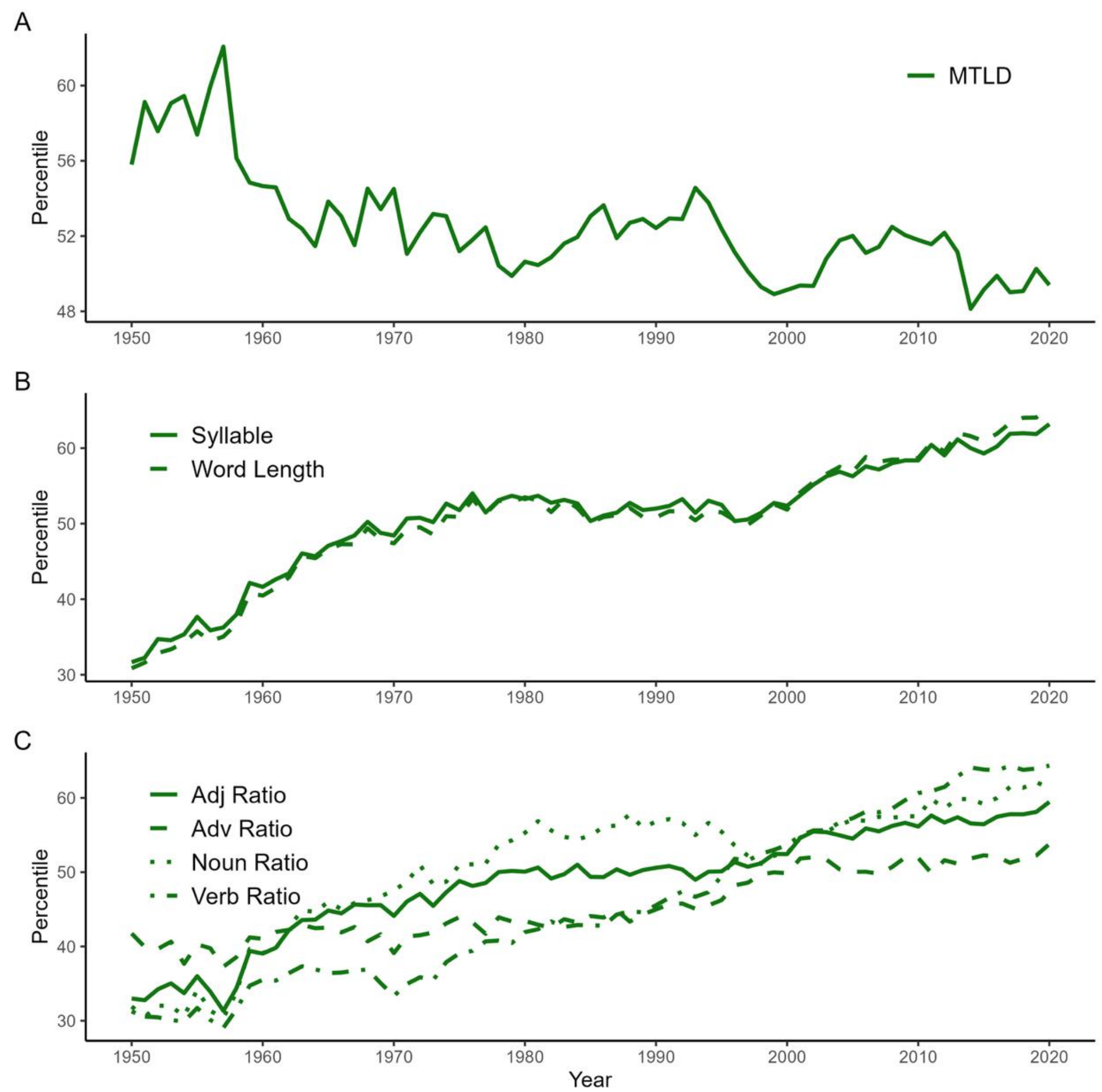

Fig S1: Historical Trajectories of Lexical Complexity Metrics in Nature, Science, and PNAS Abstracts (1950–2020). (A) Lexical diversity (MTLD), (B) Lexical sophistication (Word Length and Syllables per Word), and (C) Lexical density (Noun/Verb/Adjective/Adverb Ratios) percentile trends. Data derived from 375,000 article abstracts. Percentiles calculated within the journal-specific corpus.

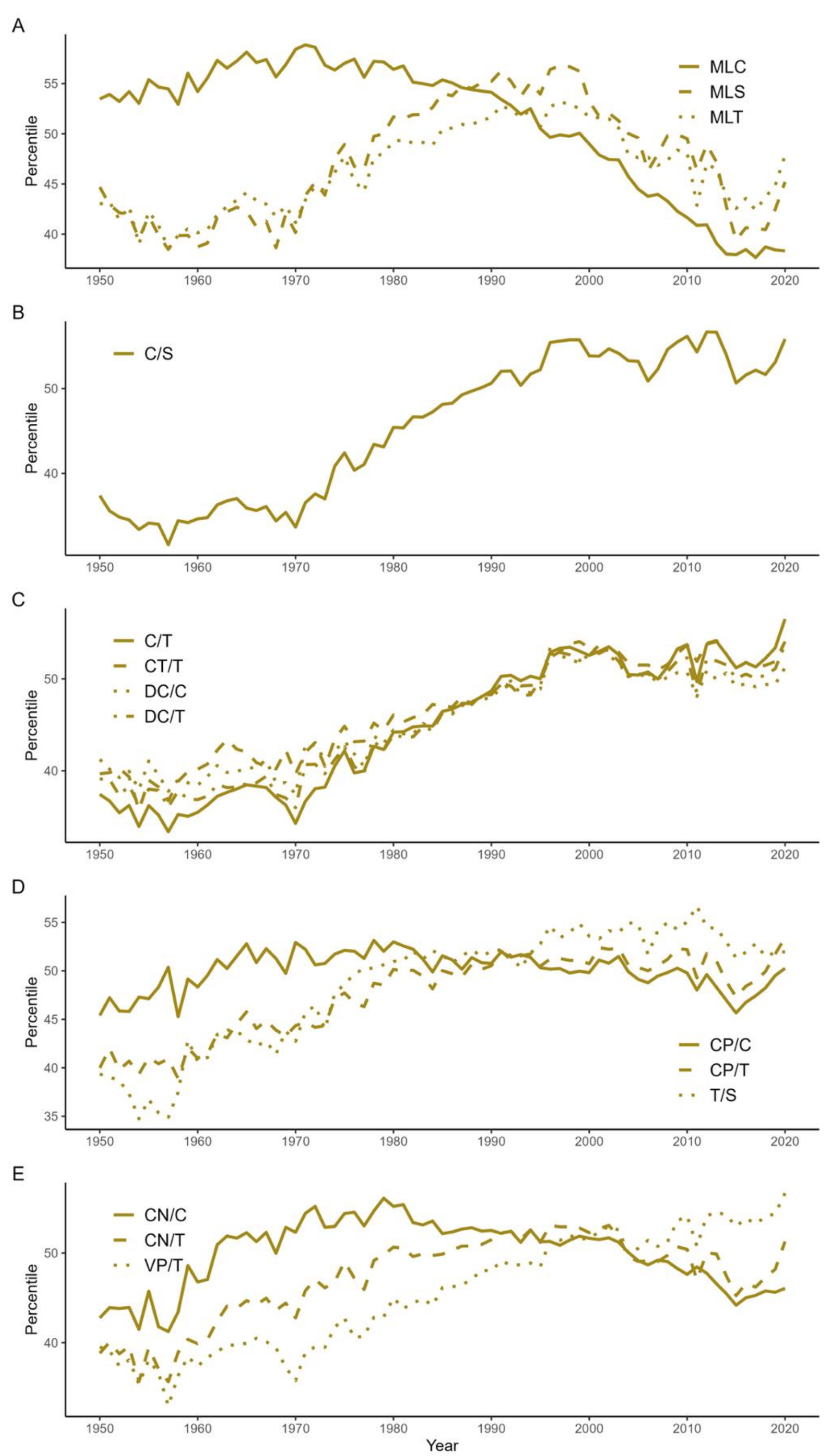

Fig S2: Syntactic Complexity Dynamics in Nature, Science, and PNAS Abstracts (1950–2020). (A) Length of production unit (MLC, MLS, MLT), (B) Sentence complexity (C/S), (C) Subordination (C/T, CT/T, DC/C, DC/T), (D) Coordination (CP/C, CP/T, T/S), and (E) Particular structures (CN/C, CN/T, VP/T) percentile trends. Data derived from 375,000 article abstracts. Percentiles calculated within the journal-specific corpus.

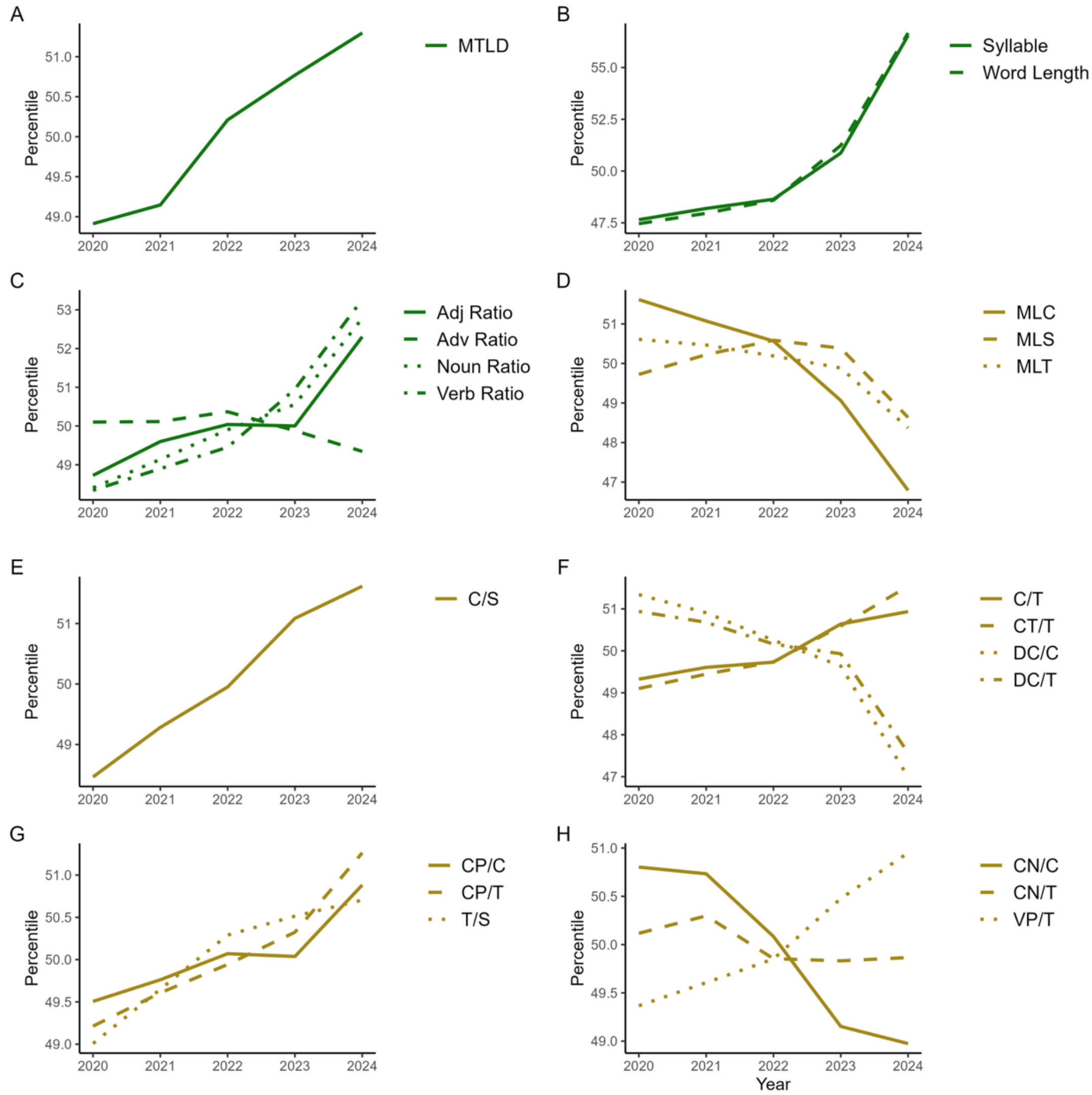

Fig S3: Temporal Evolution of Lexical and Syntactic Complexity Metrics (2020–2024). (A) Lexical diversity (MTLD), (B) Lexical sophistication (Word length, Syllables per Word), (C) Lexical density (Noun/Verb/Adjective/Adverb Ratios), (D) Length of production unit (MLC, MLS, MLT), (E) Subordination (C/T, CT/T, DC/C, DC/T), (F) Coordination (CP/C, CP/T, T/S), (G) Length of production units (MLC, MLS, MLT), (H) Particular structures (CN/C, CN/T, VP/T). Data represent percentile trends across 21.36 million scientific abstracts. Metrics calculated within annual cohorts (2020–2024) with all percentiles benchmarked against the full 5-year corpus.

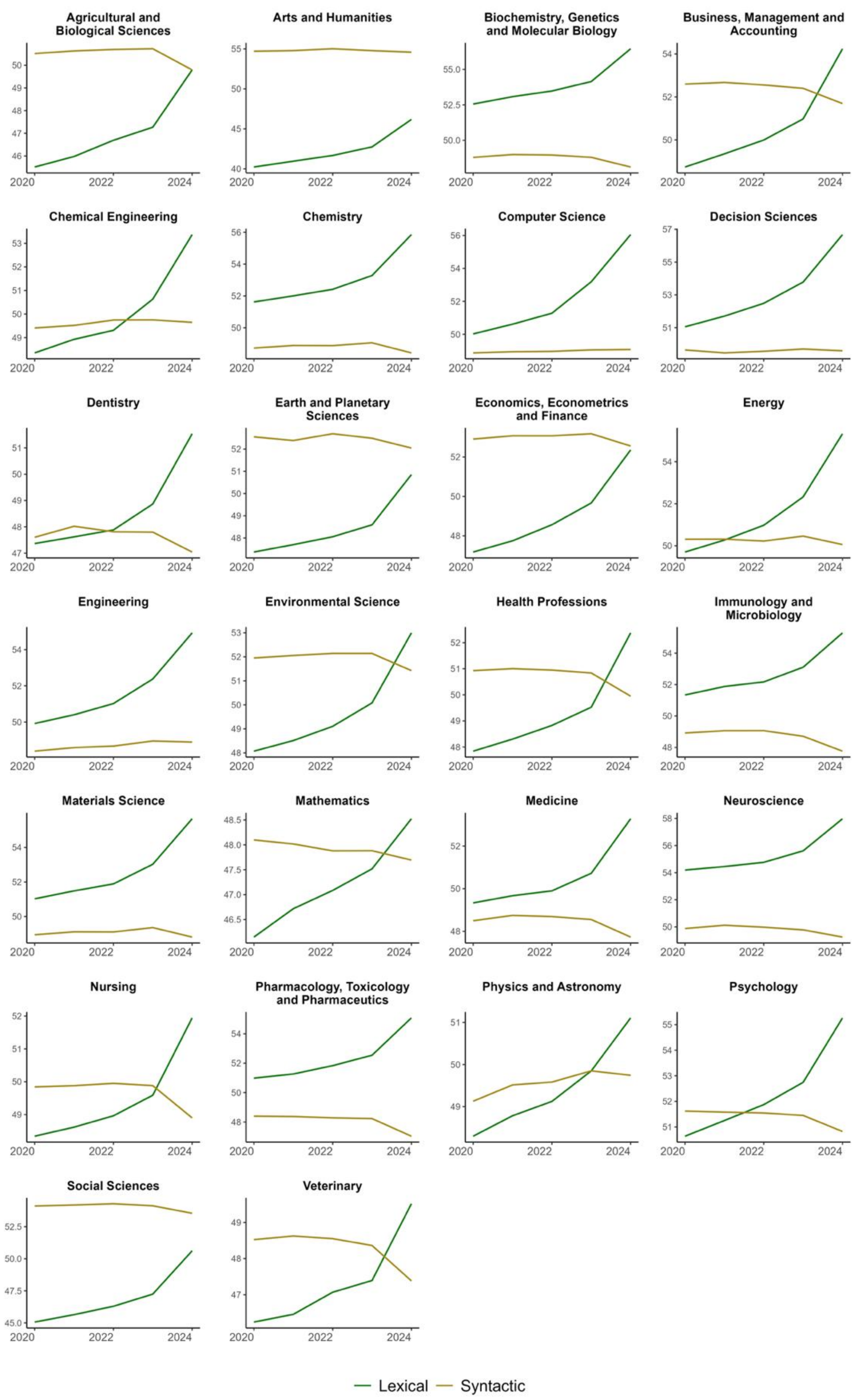

Fig S4: Field-Specific Trajectories of Lexical and Syntactic Complexity (2020–2024). Field-stratified trends in lexical (green) and syntactic (gold) complexity percentiles across 26 academic disciplines. Data derived from 21.36 million scientific abstracts, with percentiles calculated relative to the full corpus.

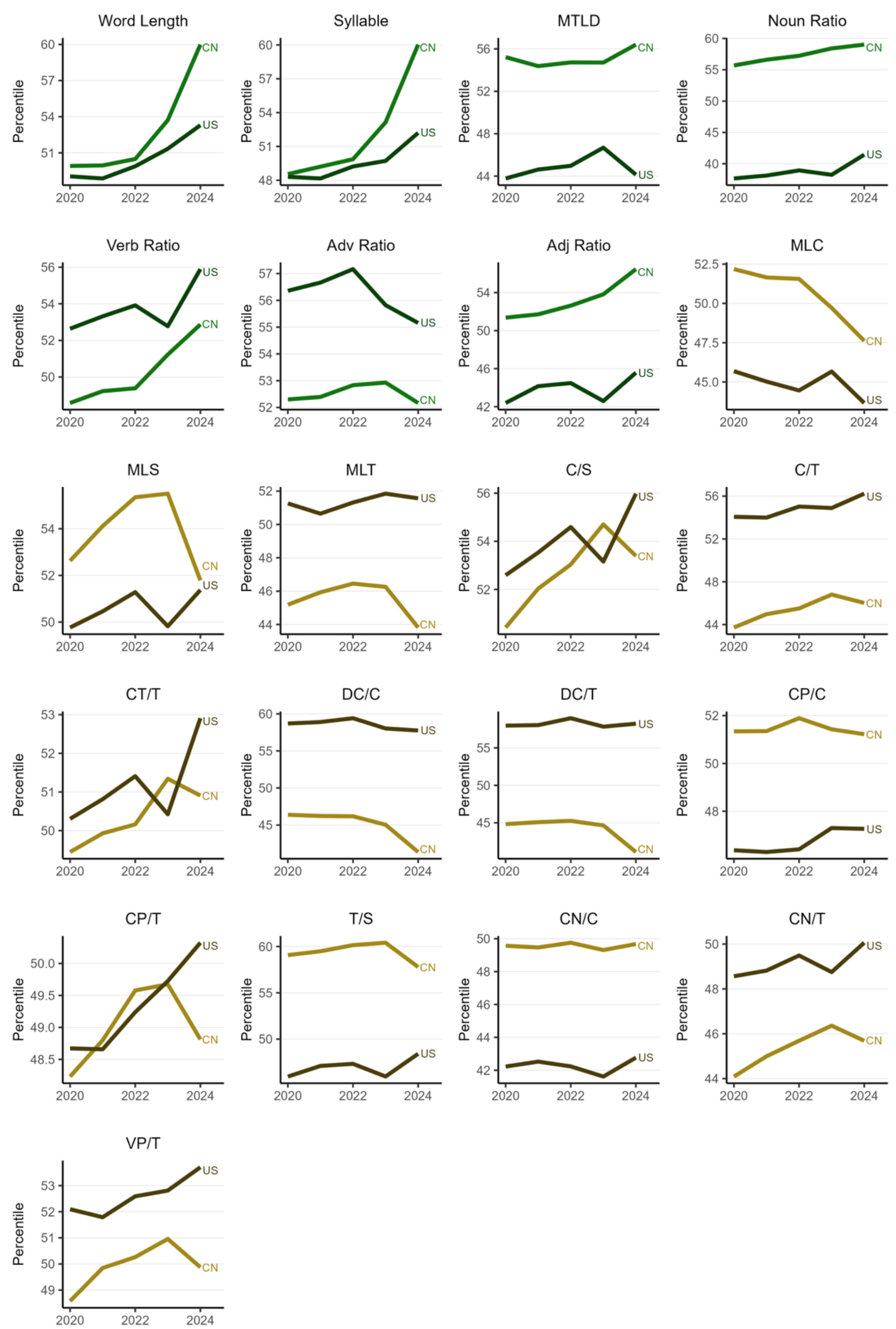

Figure S5: Annual Percentile Trends in Linguistic Complexity: Chinese (CN) vs. U.S. (US) Scholars (2020–2024). Treatment group (CN): 2.47M articles authored exclusively by name-inferred Chinese researchers affiliated with Chinese institutions. Control group (US): 331K articles authored exclusively by name-inferred English researchers affiliated with U.S. institutions.

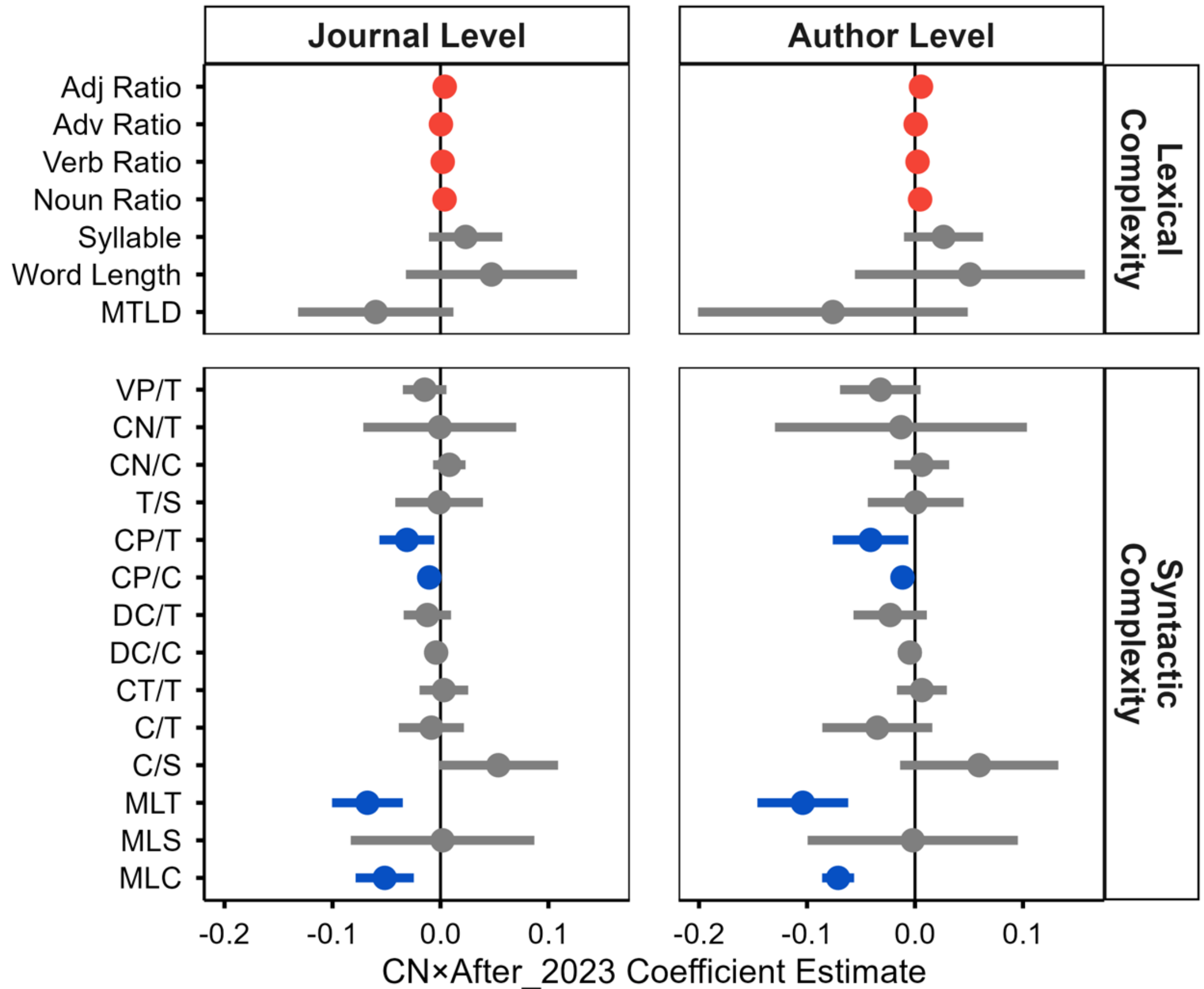

Figure S6: Coefficient Estimates for Alternative Treatment Timing CN×After_2023 interaction term across all 21 complexity metrics (row) and analytical levels (column). CN is the treatment group indicator, After_2023≥2023 (0 otherwise). Red indicates positive coefficient with $p<0.05$ significance, blue indicates negative coefficient with $p<0.05$ significance, Gray indicates statistically non-significant.

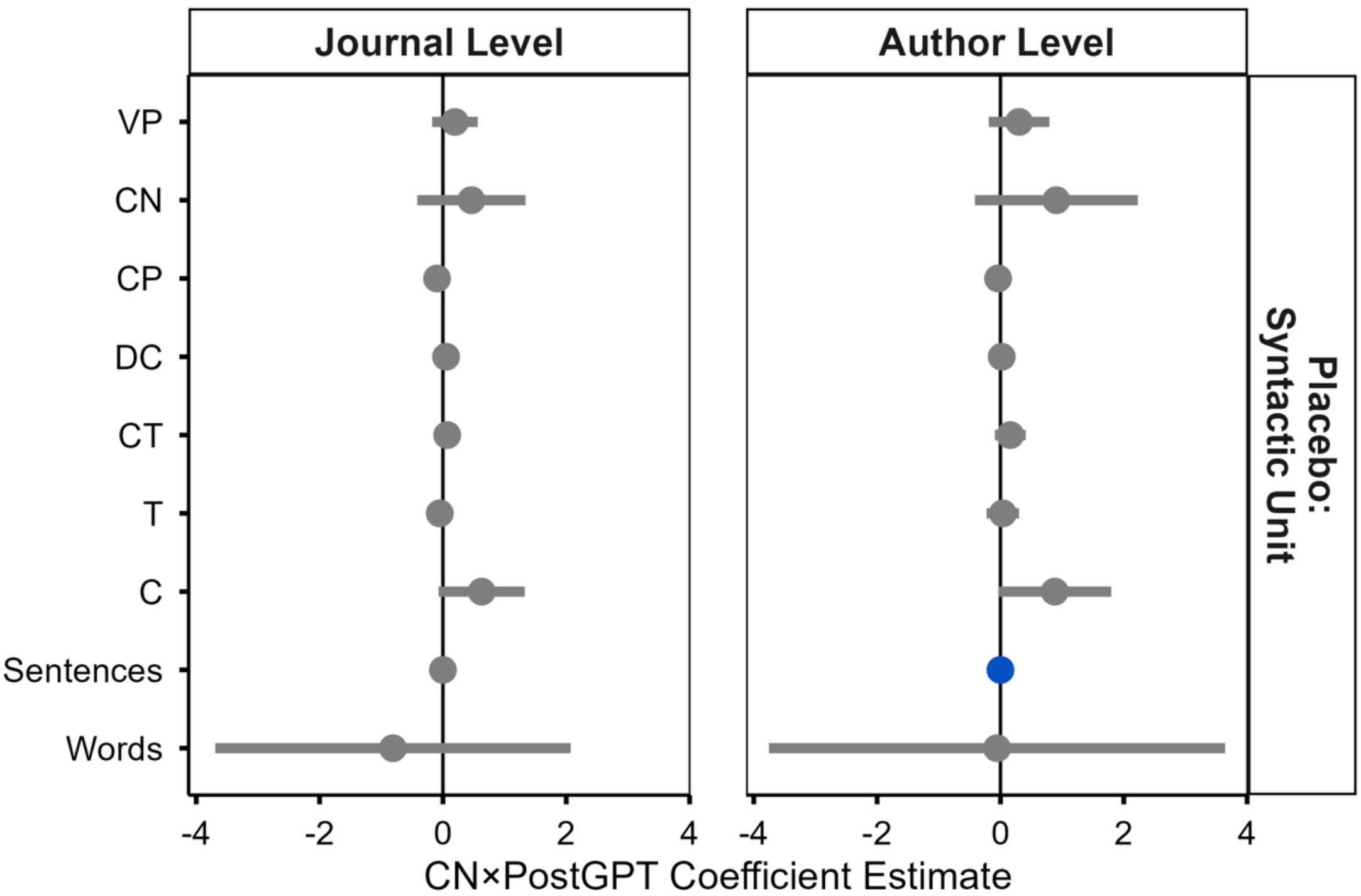

Figure S7: Coefficient estimates for CN×PostGPT interaction term across 9 basic grammatical structures (row) and analytical levels (column). CN is the treatment group indicator, PostGPT = 2024. Red indicates positive coefficient with $p<0.05$ significance, blue indicates negative coefficient with $p<0.05$ significance, Gray indicates statistically non-significant.

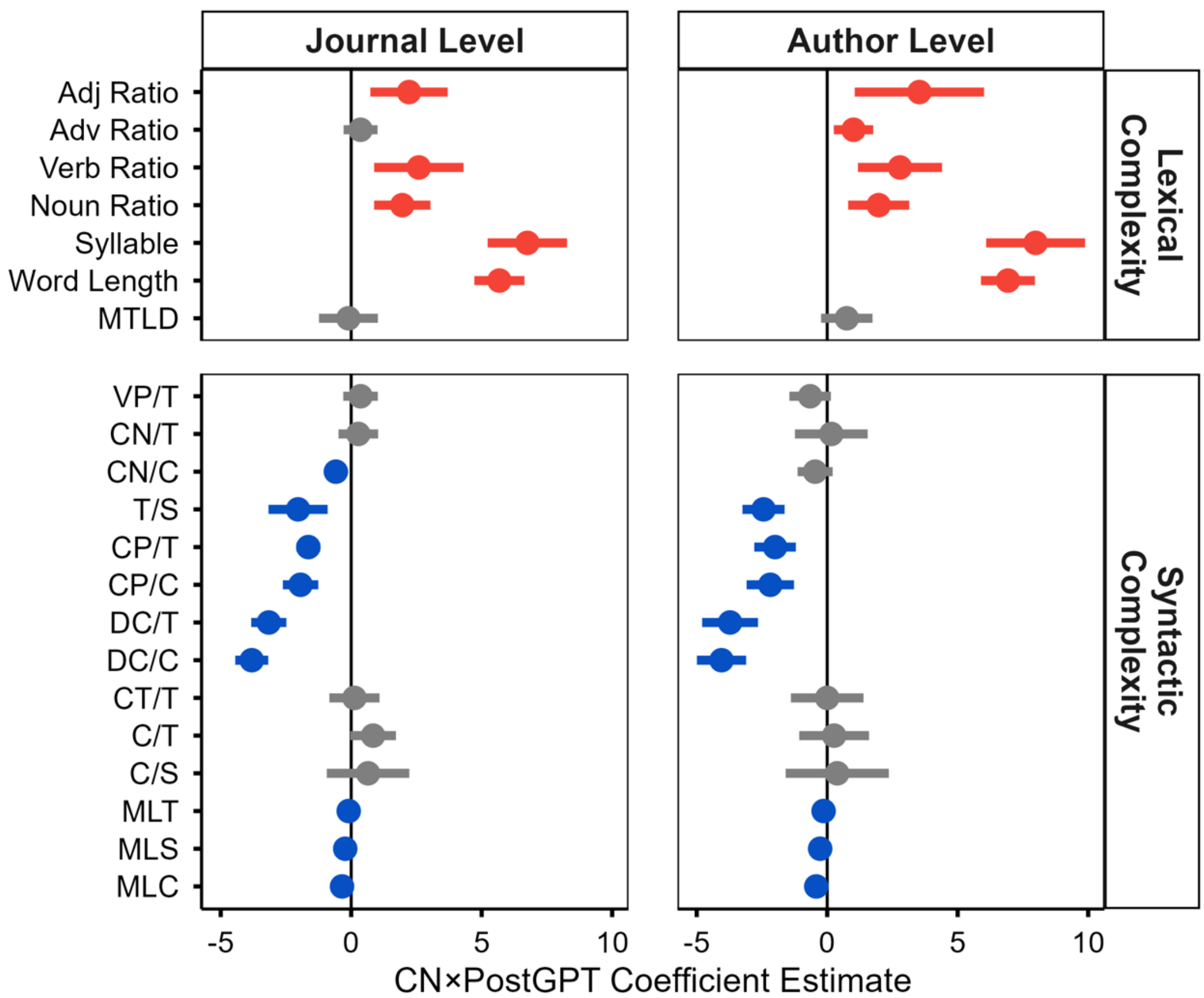

Figure S8: DID Coefficients Using Percentile Ranks as Outcome Variables. Coefficient estimates for CN×PostGPT interaction term across 21 linguistic complexity percentile ranks (row) and analytical levels (column). CN is the treatment group indicator, PostGPT = 2024. Red indicates positive coefficient with $p<0.05$ significance, blue indicates negative coefficient with $p<0.05$ significance, Gray indicates statistically non-significant.

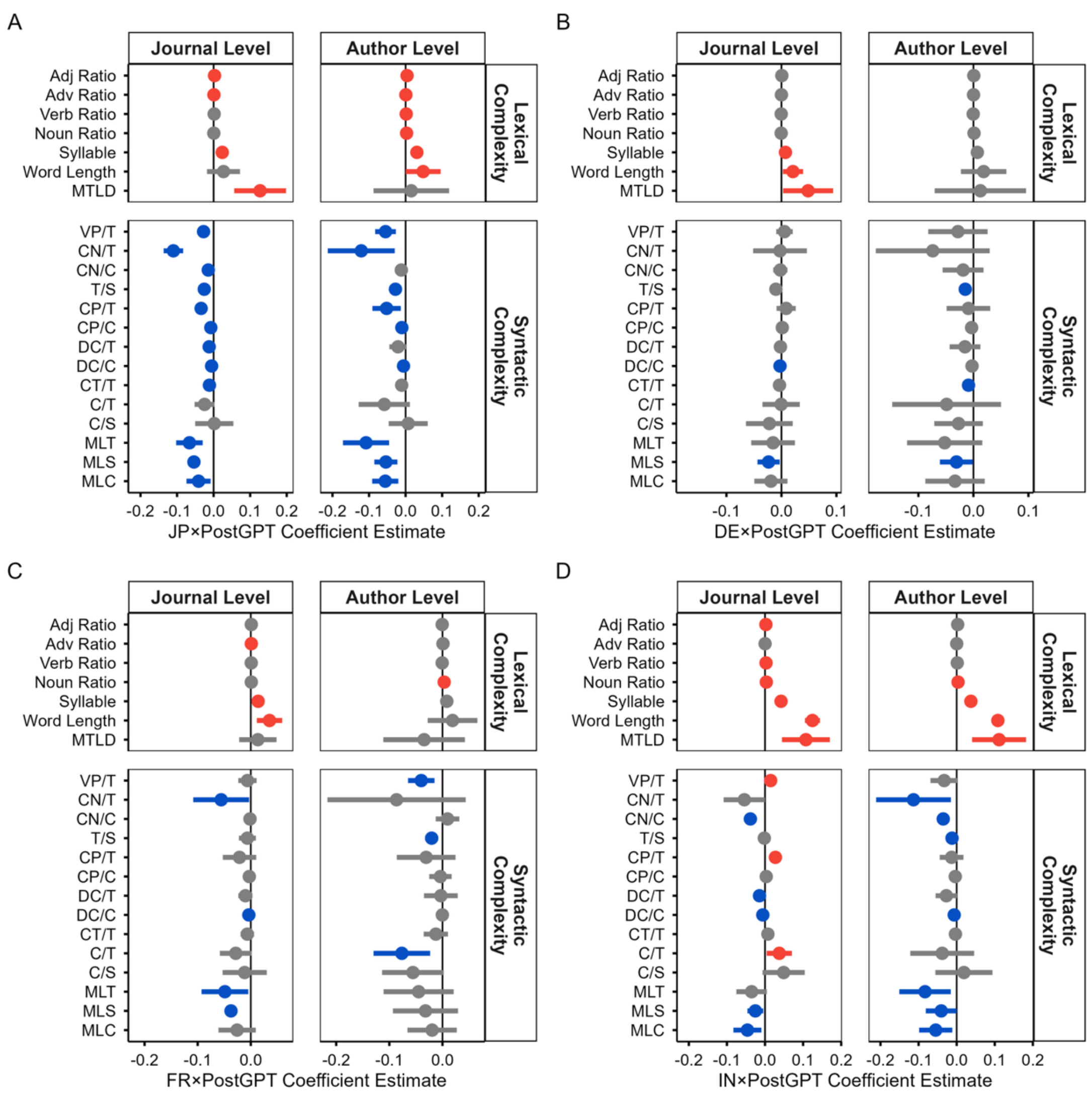

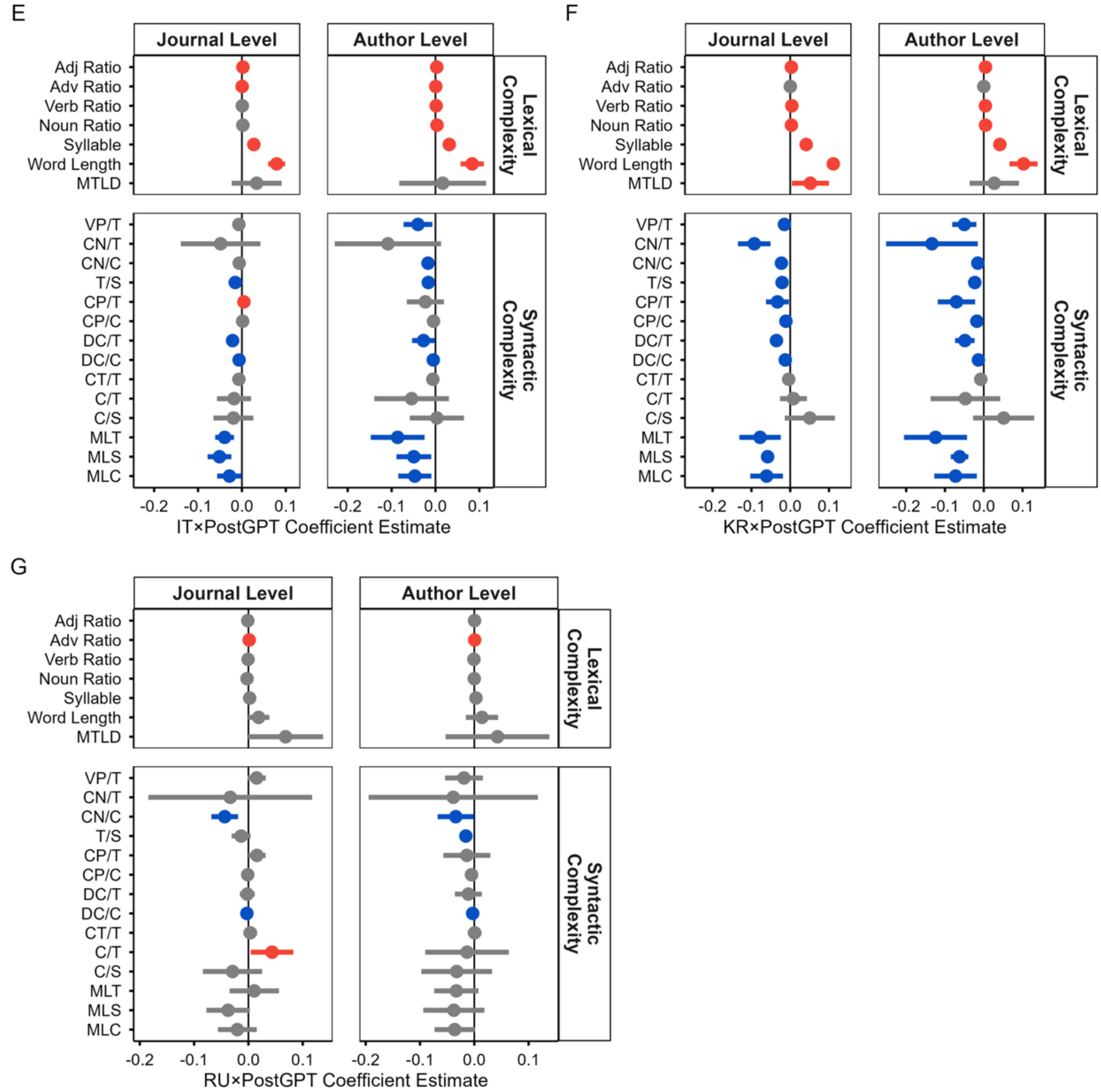

Figure S9: Heterogeneous Effects of ChatGPT Adoption Across Author Groups. Coefficient estimates for the interaction term (Author group×PostGPT) using homogeneous author-group cohorts. (A) name-inferred Japanese authors (JP), (B) name-inferred German authors (DE), (C) name-inferred French authors (FR), (D) name-inferred Indian authors (IN), (E) name-inferred Italian authors (IT), (F) name-inferred Korean authors (KR) , (G) name-inferred Russian authors (RU). PostGPT = 2024. Red indicates positive coefficient with $p<0.05$ significance, blue indicates negative coefficient with $p<0.05$ significance, Gray indicates statistically non-significant.

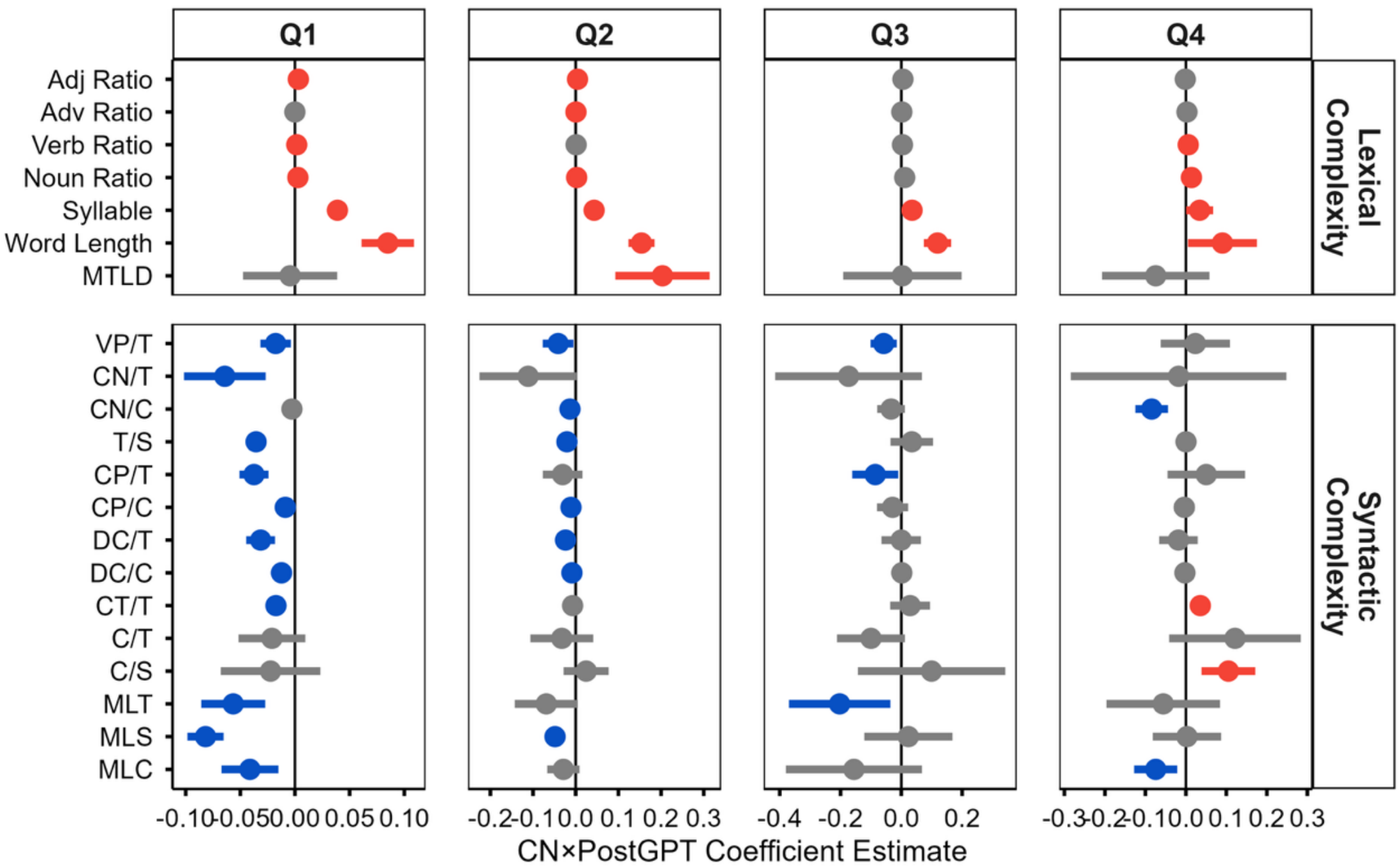

Figure S10: Journal Impact Factor Quartile-Stratified Effects of ChatGPT on Linguistic Complexity. Coefficient Estimates for CN×PostGPT interaction term across all 21 complexity metrics (row) and Quartile-Stratified subsamples (column). CN is the treatment group indicator, PostGPT = 2024. Red indicates positive coefficient with $p<0.05$ significance, blue indicates negative coefficient with $p<0.05$ significance, Gray indicates statistically non-significant.

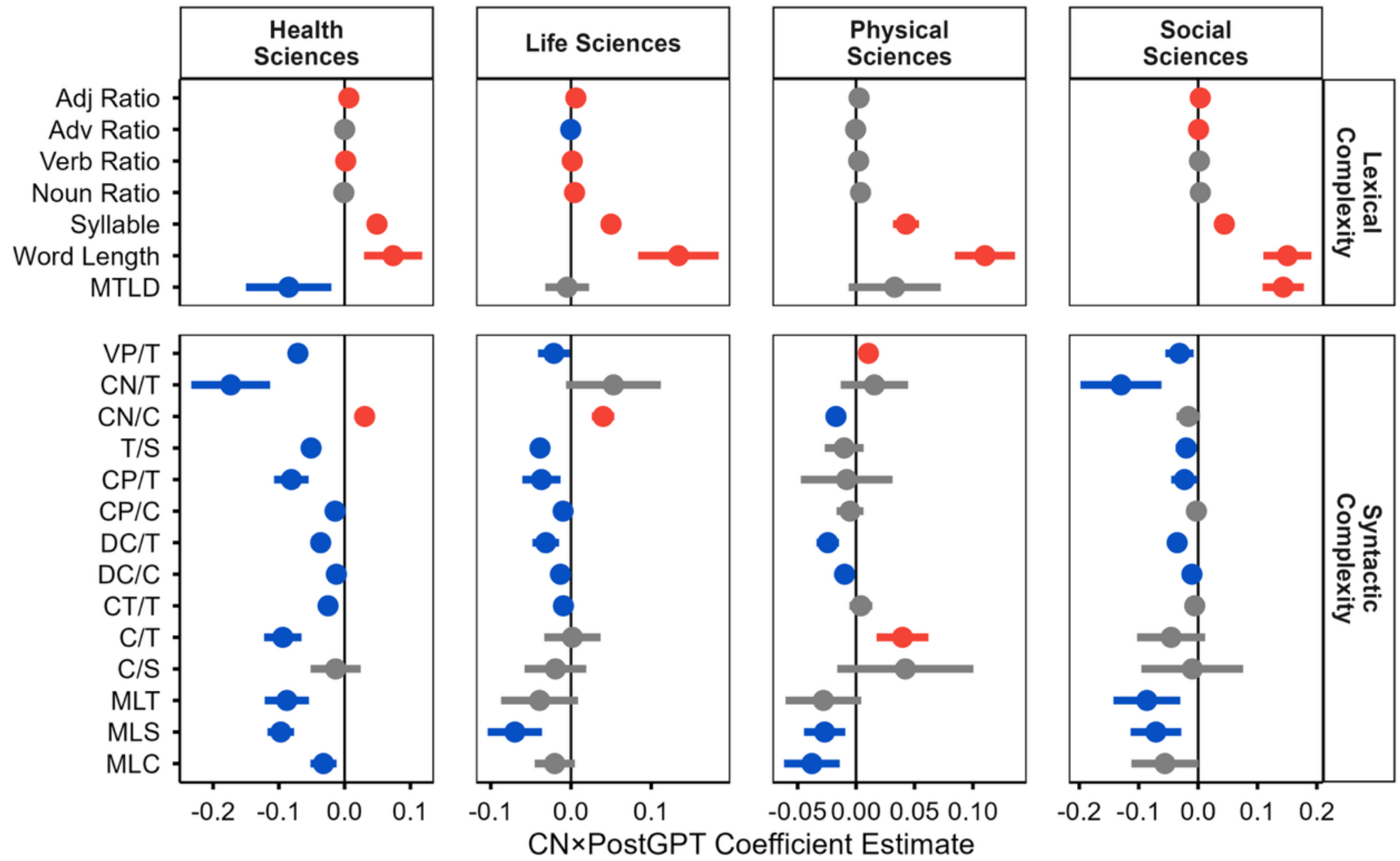

Figure S11: Domain-Stratified Effects of ChatGPT on Linguistic Complexity. Coefficient Estimates for CN×PostGPT interaction term across all 21 complexity metrics (row) and Domain-Stratified subsamples (column). CN is the treatment group indicator, PostGPT = 2024. Red indicates positive coefficient with $p<0.05$ significance, blue indicates negative coefficient with $p<0.05$ significance, Gray indicates statistically non-significant.

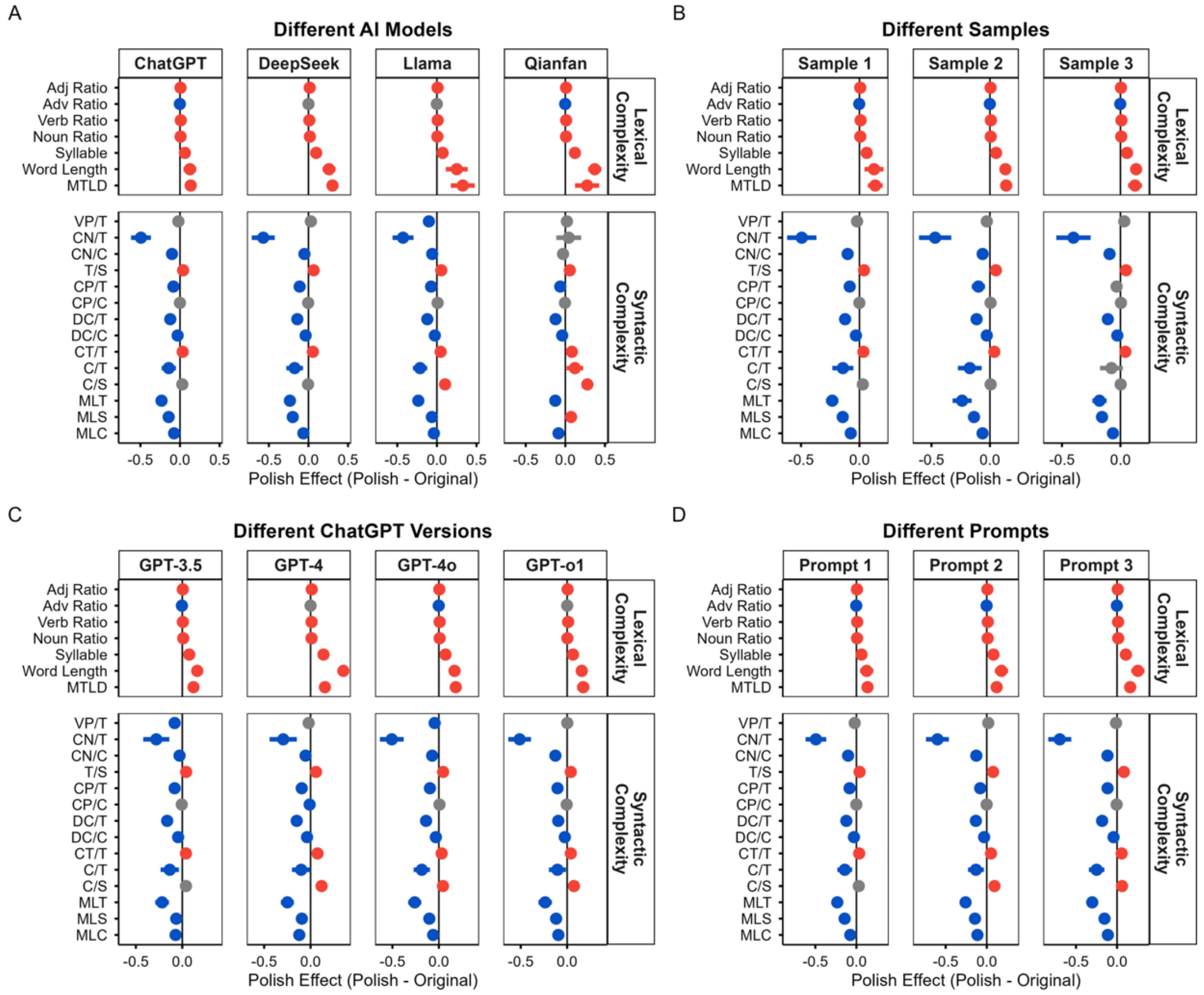

Figure S12: Experimental Validation of LLM Polishing Effects on Linguistic Complexity. We use randomly sampled abstracts from English NES samples to polish by large language models, then quantify linguistic complexity changes via paired t-tests. (A) Cross-Model Comparison. Paired t-test results comparing complexity metrics before vs. after polishing identical text samples by four LLMs: ChatGPT, DeepSeek, LLaMA, and Qianwen. (B) Cross-Sample Generalization (ChatGPT). Paired t-test results for three distinct text samples polished by ChatGPT. (C) Version-Specific Effects (ChatGPT). Paired t-test results for Identical sample polished by different ChatGPT iterations: v3.5, v4, v4-o, and v-o1. (D) Prompt Engineering Impacts. Identical sample polished by ChatGPT using three distinct instructional prompts. Red indicates positive coefficient with $p<0.05$ significance, blue indicates negative coefficient with $p<0.05$ significance, Gray indicates statistically non-significant.

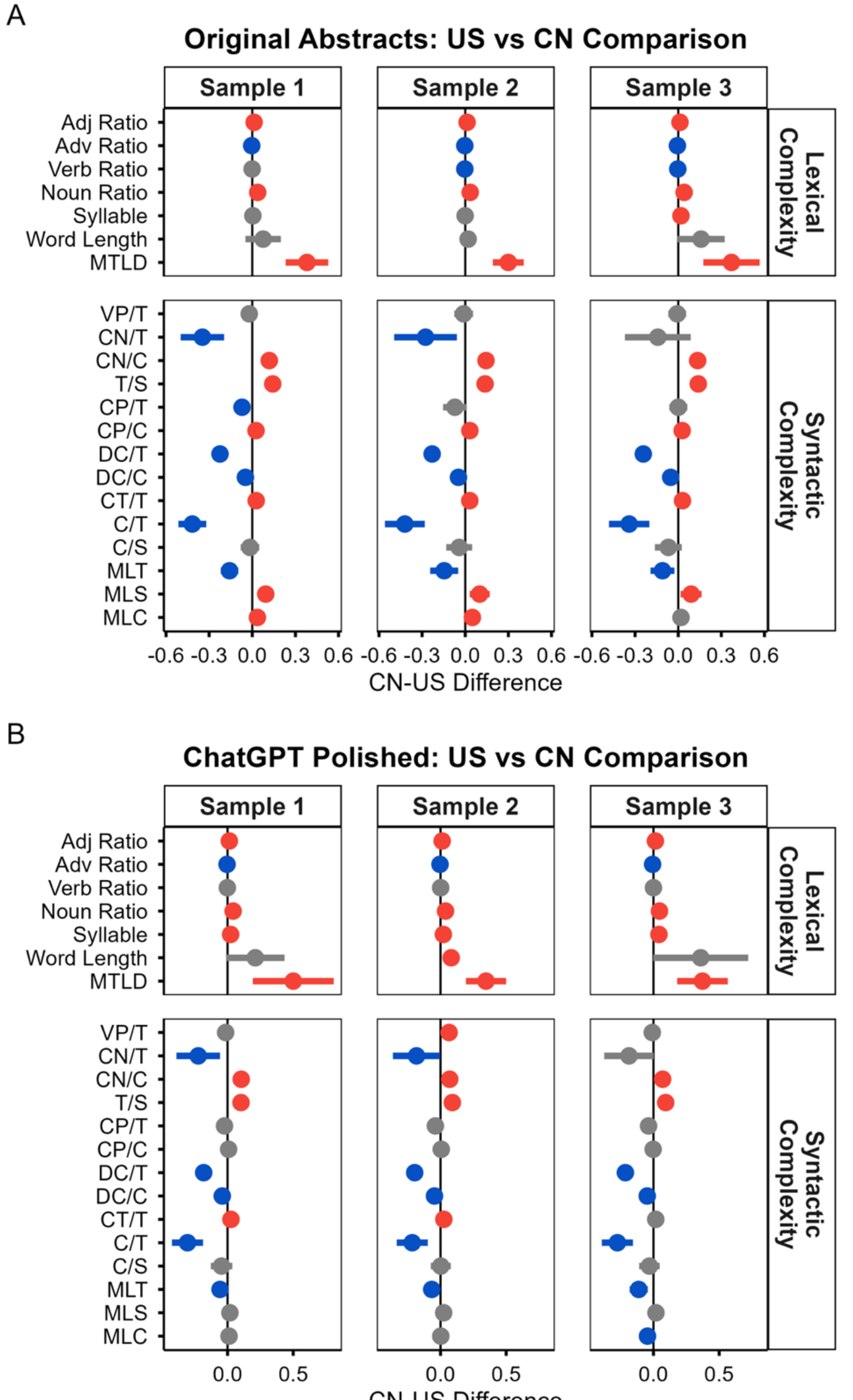

Figure S13: Persistent Linguistic Differences: Cross-Group Comparison Before and After ChatGPT Polishing. (A) Original abstract differences: Independent t-test results comparing CN and US sample linguistic complexity across three 1,000-abstract samples. (B) Post-polishing differences: Complexity differentials after ChatGPT-4o (Prompt1) processing. Red indicates positive coefficient with $p<0.05$ significance, blue indicates negative coefficient with $p<0.05$ significance, Gray indicates statistically non-significant.

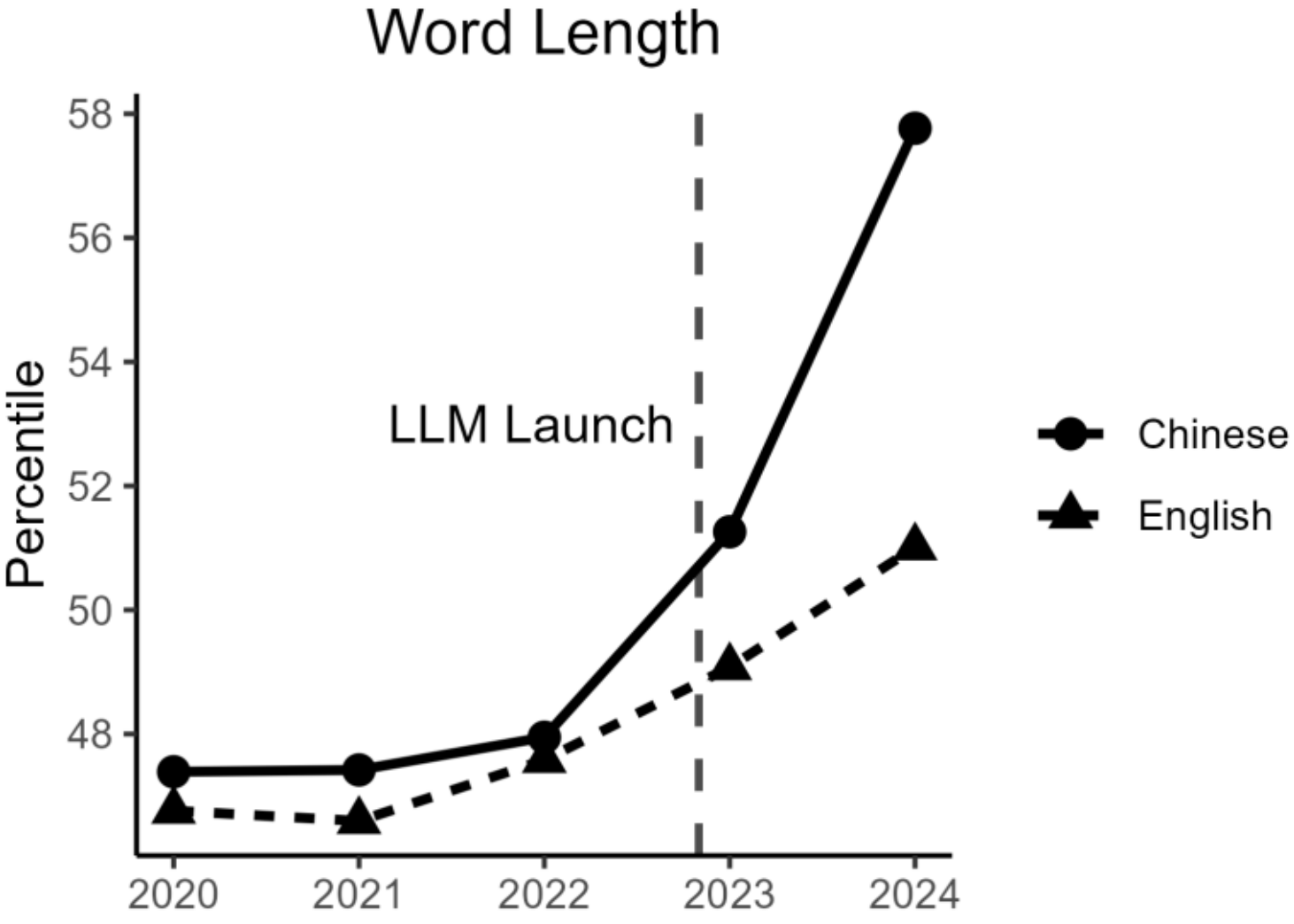

Figure S14. Mean Word Length Percentiles by Group and Publication Year (2020–2024).

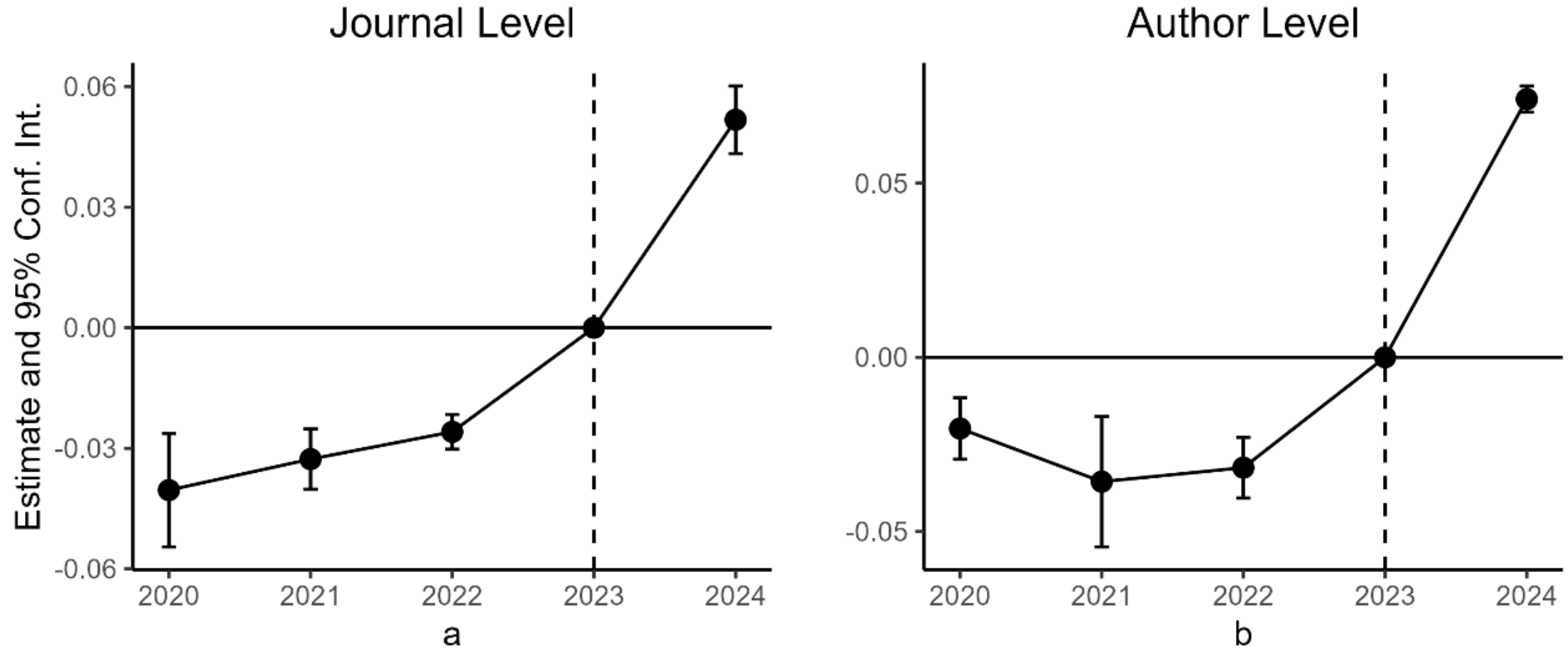

Figure S15. Event Study of ChatGPT's Impact on Non-Native Academic Writing (2020–2024) at (a) Journal level and (b) Author level.

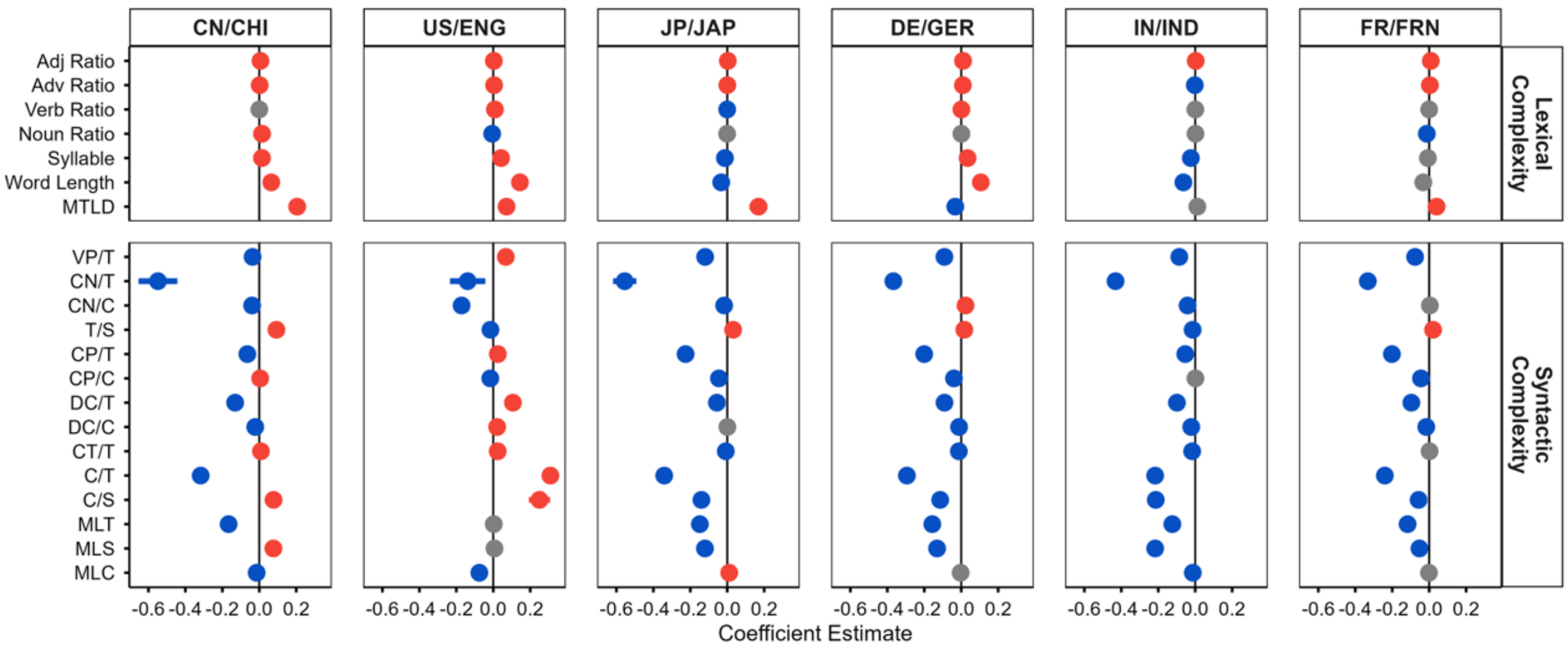

Figure S16: Cross-Group Linguistic Signatures in Academic Writing: Coefficient Estimates of Syntactic and Lexical Feature Preferences. OLS regression analysis of author-group-linked linguistic patterns across 21.36M publications (2020–2024), controlling for citations, authorship size, open-access status, sentence count, and year fixed effects. Coefficient estimates reflect deviations from global annual averages. Red indicates positive coefficient with $p<0.05$ significance, blue indicates negative coefficient with $p<0.05$ significance, Gray indicates statistically non-significant.

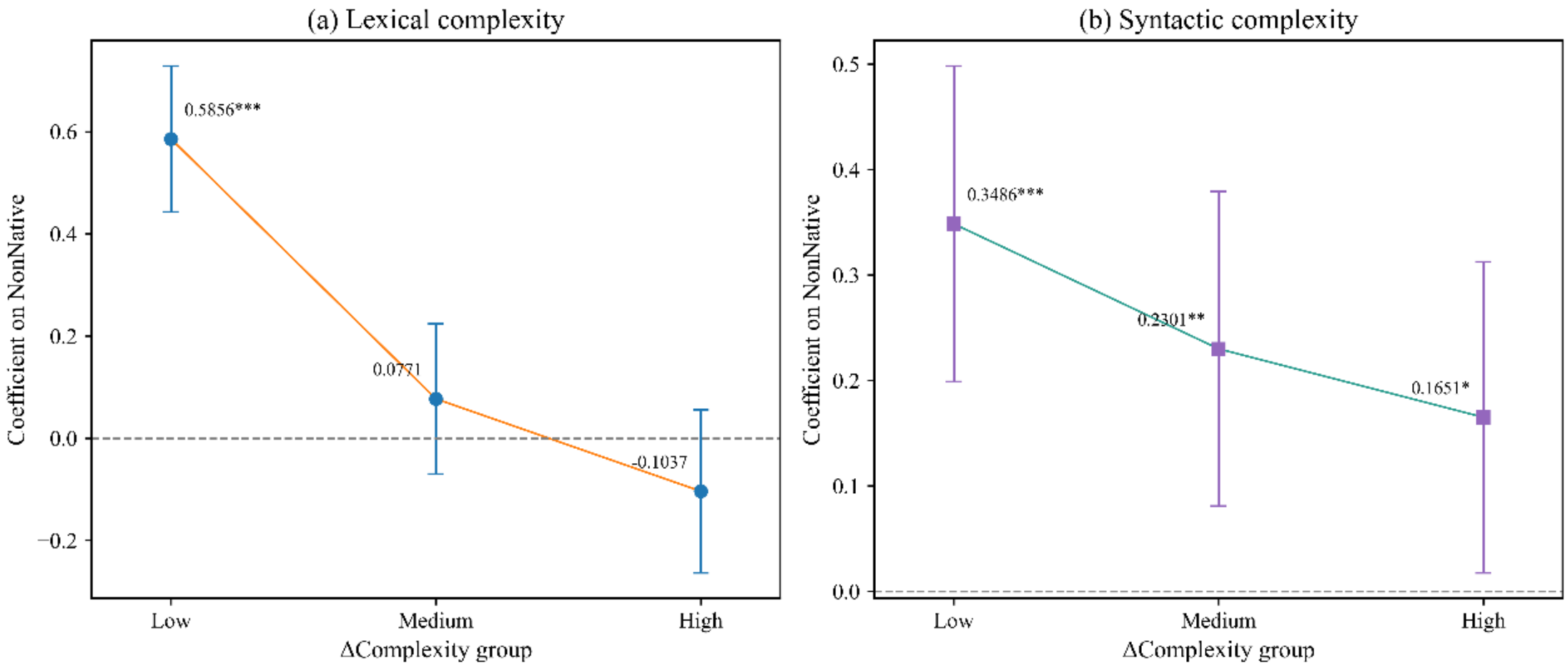

Figure S17. Grouped regression results using changes in publication rank as the dependent variable.

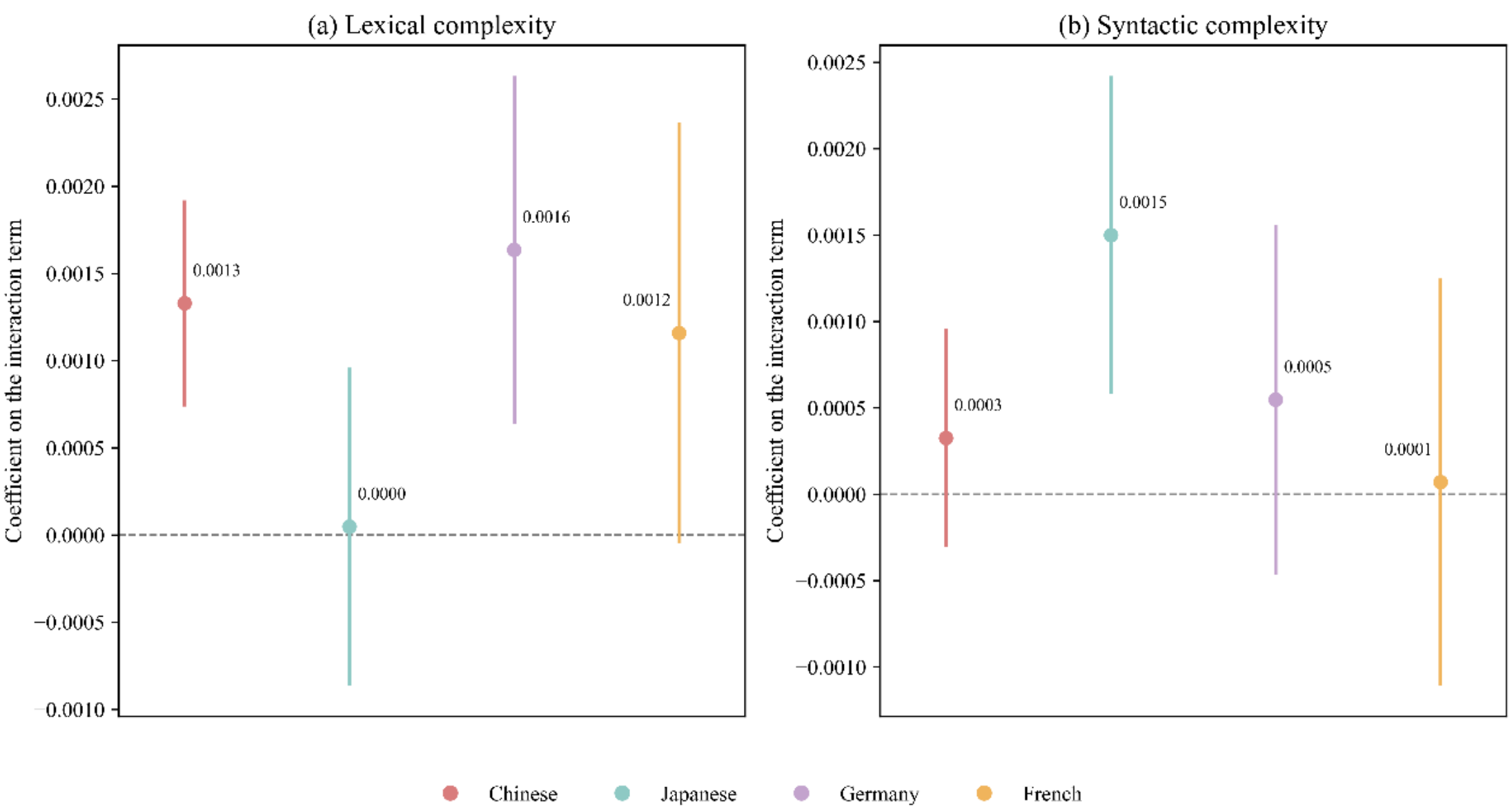

Figure S18. Cross-country heterogeneity in the effect of language complexity on NNES-NES publication output differences.

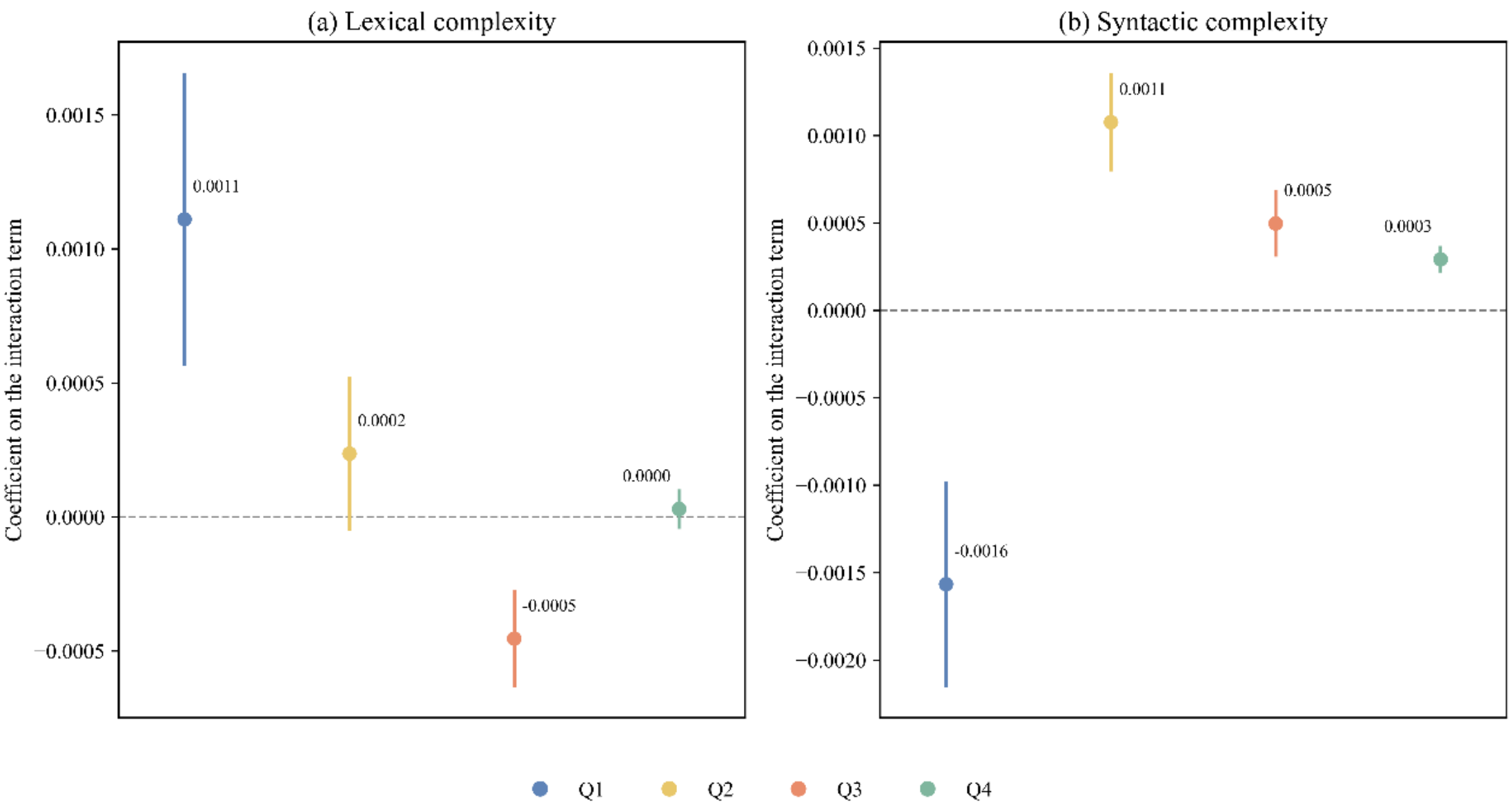

Figure S19. Heterogeneous effects across journal quartiles (Q1-Q4).

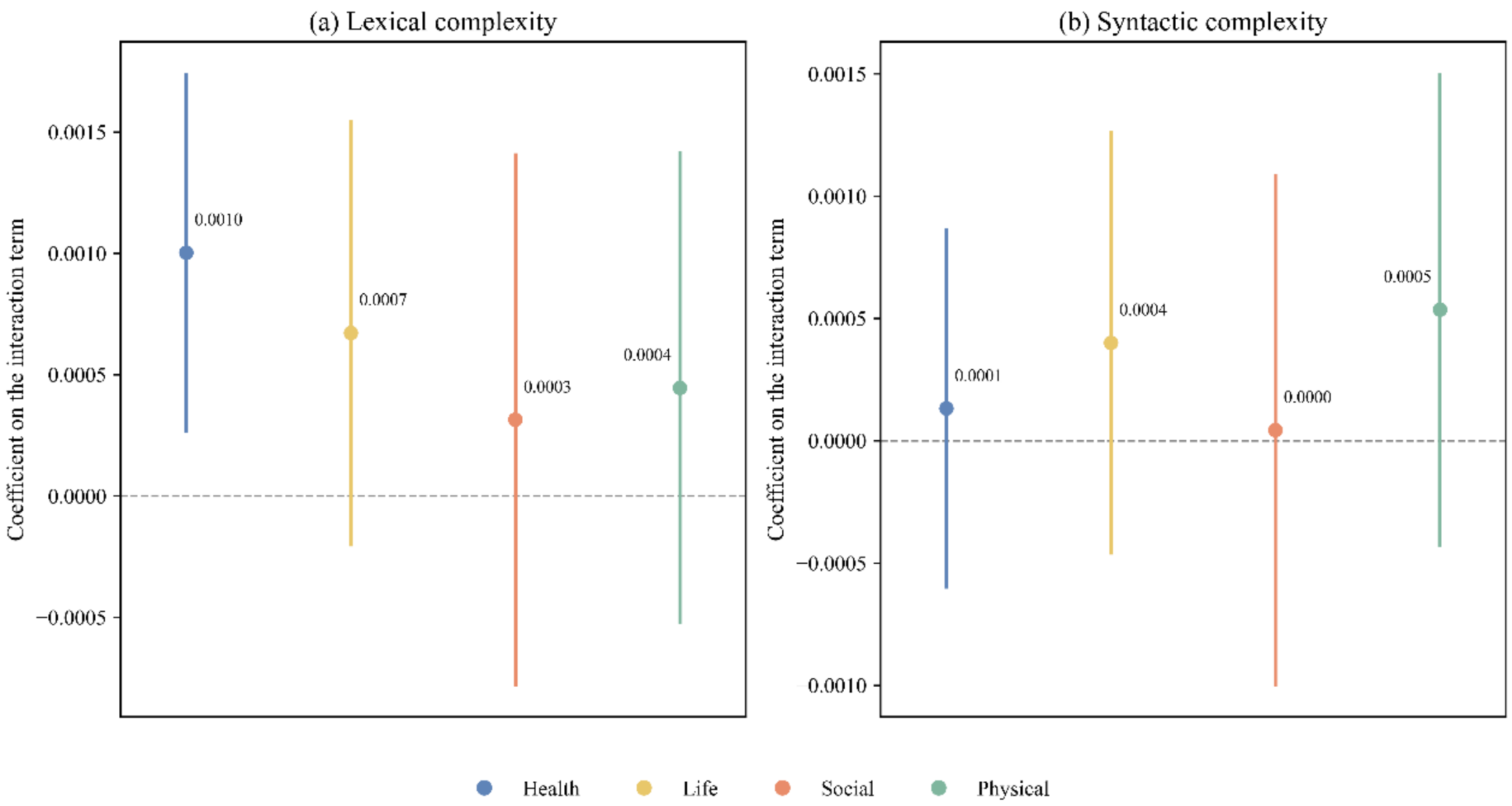

Figure S20. Field-level heterogeneity in the effect of language complexity on publication productivity.

# S8 Supplementary Information Tables

| Dimension | Category | Indicator | Definition |
|---|---|---|---|
| lexical complexity | lexical diversity | MTLD | An indicator used to measure the lexical diversity of text. It avoids the stability of lexical diversity measurement affected by text length by maintaining the TTR (Type-Token Ratio) of the text at around 0.72. |
| | lexical sophistication | Word Length | The average number of characters in the words of the text. |
| | | Syllable | The average number of syllables in the words of the text. |
| | lexical density | Noun Ratio | The proportion of Nouns in all words of the text. |
| | | Adj Ratio | The proportion of Adjectives in all words of the text. |
| | | Adv Ratio | The proportion of Adverb in all words of the text. |
| | | Verb Ratio | The proportion of Verb in all words of the text. |
| Syntactic complexity | Length of production unit | MLC | Mean length of clause (# of words / # of clauses) |
| | | MLS | Mean length of sentence (# of words / # of sentences) |
| | | MLT | Mean length of T-unit (# of words / # of T-units) |
| | Sentence complexity | C/S | Sentence complexity ratio (# of clauses / # of sentences) |
| | Subordination | C/T | T-unit complexity ratio (# of clauses / # of T-units) |
| | | CT/T | Complex T-unit ratio (# of complex T-units / # of T-units) |
| | | DC/C | Dependent clause ratio (# of dependent clauses / # of clauses) |
| | | DC/T | Dependent clauses per T-unit (# of dependent clauses / # of T-units) |
| | Coordination | CP/C | Coordinate phrases per clause (# of coordinate phrases / # of clauses) |
| | | CP/T | Coordinate phrases per T-unit (# of coordinate phrases / # of T-units) |
| | | T/S | Sentence coordination ratio (# of T-units / # of sentences) |
| | Particular structures | CN/C | Complex nominals per clause (# of complex nominals / # of clauses) |
| | | CN/T | Complex nominals per T-unit (# of complex nominals / # of T-units) |
| | | VP/S | Verb phrases per T-unit (# of verb phrases / # of T-units) |

Table S1. Multidimensional Linguistic Complexity Indicator System.

| field_name | domain_name |
|---|---|
| Medicine | Health Sciences |
| Dentistry | Health Sciences |
| Health Professions | Health Sciences |
| Nursing | Health Sciences |
| Veterinary | Health Sciences |
| Agricultural and Biological Sciences | Life Sciences |
| Biochemistry, Genetics and Molecular Biology | Life Sciences |
| Immunology and Microbiology | Life Sciences |
| Neuroscience | Life Sciences |
| Pharmacology, Toxicology and Pharmaceutics | Life Sciences |
| Earth and Planetary Sciences | Physical Sciences |
| Physics and Astronomy | Physical Sciences |
| Environmental Science | Physical Sciences |
| Chemistry | Physical Sciences |
| Engineering | Physical Sciences |
| Computer Science | Physical Sciences |
| Energy | Physical Sciences |
| Mathematics | Physical Sciences |
| Materials Science | Physical Sciences |
| Chemical Engineering | Physical Sciences |
| Business, Management and Accounting | Social Sciences |
| Economics, Econometrics and Finance | Social Sciences |
| Social Sciences | Social Sciences |
| Arts and Humanities | Social Sciences |
| Decision Sciences | Social Sciences |
| Psychology | Social Sciences |

Table S2: Domain-Field Mapping in OpenAlex Taxonomy.

| Code | Country Name |
| --- | --- |
| FR | France |
| CN | China |
| DE | Germany |
| IN | India |
| JP | Japan |
| US | United States |
| GB | United Kingdom |
| KR | South Korea |
| IT | Italy |
| RU | Russia |

Table S3: ISO 3166-1 Alpha-2 Country Codes for Nations Referenced in This Study.

| Ethnicity | Wikipedia Categories |
| --- | --- |
| MEA | Egyptian, Iraqi, Iranian, Lebanese, Syrian, Tunisian |
| IND | Indian |
| ENG | British |
| FRN | French |
| GER | German |
| ITA | Italian |
| SPA | Columbian, Spanish, Venezuelan |
| RUS | Russian |
| CHI | Chinese |
| JAP | Japanese |
| KOR | Korean |
| VIE | Vietnamese |

Table S4: Name-Inferred Author Group Classification Schema in Ethnicseer.

| measure | count | mean | std | 25% | 50% | 75% |
|---|---|---|---|---|---|---|
| authors_count | 2470714 | 5.53 | 3.00 | 3 | 5 | 7 |
| year | 2470714 | 2022.06 | 1.33 | 2021 | 2022 | 2023 |
| citation | 2470714 | 8.02 | 50.46 | 0 | 2 | 8 |
| nwords | 2470714 | 200.82 | 151.60 | 149 | 194 | 241 |
| sentence | 2470714 | 8.20 | 7.34 | 6 | 8 | 10 |
| health | 2462674 | 0.23 | 0.42 | 0 | 0 | 0 |
| life | 2462674 | 0.24 | 0.43 | 0 | 0 | 0 |
| social | 2462674 | 0.11 | 0.32 | 0 | 0 | 0 |
| physical | 2462674 | 0.69 | 0.46 | 0 | 1 | 1 |
| quantile | 2268635 | 1.25 | 0.60 | 1 | 1 | 1 |
| after | 2470714 | 0.17 | 0.37 | 0 | 0 | 0 |
| mtld | 2470714 | 15.08 | 2.92 | 14.42 | 14.90 | 15.47 |
| syllable | 2470714 | 1.88 | 0.17 | 1.77 | 1.87 | 1.98 |
| wrd_length | 2470714 | 5.67 | 2.32 | 5.32 | 5.59 | 5.87 |
| noun_ratio | 2470714 | 0.32 | 0.05 | 0.29 | 0.32 | 0.35 |
| verb_ratio | 2470714 | 0.10 | 0.03 | 0.09 | 0.10 | 0.12 |
| adv_ratio | 2470714 | 0.03 | 0.02 | 0.02 | 0.03 | 0.04 |
| adj_ratio | 2470714 | 0.12 | 0.04 | 0.09 | 0.11 | 0.14 |
| mlc | 2470714 | 10.23 | 3.63 | 8.20 | 9.63 | 11.50 |
| mls | 2470714 | 25.64 | 7.53 | 21.22 | 24.57 | 28.67 |
| mlt | 2470714 | 30.56 | 14.55 | 23.33 | 28.00 | 34.43 |
| c_s | 2470714 | 2.65 | 0.93 | 2.00 | 2.56 | 3.11 |
| c_t | 2470714 | 3.09 | 1.27 | 2.38 | 2.89 | 3.57 |
| ct_t | 2470714 | 0.75 | 0.22 | 0.64 | 0.78 | 0.91 |
| dc_c | 2470714 | 0.11 | 0.08 | 0.05 | 0.10 | 0.16 |
| dc_t | 2470714 | 0.35 | 0.33 | 0.14 | 0.29 | 0.50 |
| cp_c | 2470714 | 0.29 | 0.21 | 0.16 | 0.25 | 0.38 |
| cp_t | 2470714 | 0.85 | 0.68 | 0.45 | 0.75 | 1.09 |
| t_s | 2470714 | 0.87 | 0.24 | 0.75 | 0.88 | 1.00 |
| cn_c | 2470714 | 1.46 | 0.55 | 1.12 | 1.37 | 1.69 |
| cn_t | 2470714 | 4.38 | 2.14 | 3.17 | 4.00 | 5.00 |
| vp_t | 2470714 | 1.75 | 0.58 | 1.43 | 1.67 | 2.00 |

Table S5: Descriptive Statistics: Treatment Group (Chinese NNES Authors).

| measure | count | mean | std | 25% | 50% | 75% |
|---|---|---|---|---|---|---|
| authors_count | 330504 | 1.74 | 1.23 | 1 | 1 | 2 |
| year | 330504 | 2021.68 | 1.36 | 2020 | 2022 | 2023 |
| citation | 330504 | 4.07 | 20.03 | 0 | 0 | 3 |
| nwords | 330504 | 207.42 | 295.91 | 101 | 161 | 237 |
| sentence | 330504 | 8.81 | 13.46 | 4 | 6 | 10 |
| health | 324554 | 0.31 | 0.46 | 0 | 0 | 1 |
| life | 324554 | 0.15 | 0.36 | 0 | 0 | 0 |
| social | 324554 | 0.59 | 0.49 | 0 | 1 | 1 |
| physical | 324554 | 0.28 | 0.45 | 0 | 0 | 1 |
| quantile | 309969 | 1.53 | 0.78 | 1 | 1 | 2 |
| after | 330504 | 0.12 | 0.32 | 0 | 0 | 0 |
| mtld | 330504 | 14.77 | 1.43 | 14.15 | 14.59 | 15.15 |
| syllable | 330504 | 1.86 | 0.20 | 1.75 | 1.86 | 1.97 |
| wrd_length | 330504 | 5.56 | 0.64 | 5.24 | 5.55 | 5.87 |
| noun_ratio | 330504 | 0.28 | 0.07 | 0.24 | 0.29 | 0.33 |
| verb_ratio | 330504 | 0.11 | 0.04 | 0.09 | 0.11 | 0.13 |
| adv_ratio | 330504 | 0.03 | 0.02 | 0.02 | 0.03 | 0.05 |
| adj_ratio | 330504 | 0.10 | 0.05 | 0.08 | 0.10 | 0.13 |
| mlc | 330504 | 10.22 | 6.65 | 7.65 | 9.10 | 11.06 |
| mls | 330504 | 24.65 | 9.27 | 19.78 | 24.22 | 29.00 |
| mlt | 330504 | 33.61 | 22.98 | 23.83 | 30.25 | 39.00 |
| c_s | 330504 | 2.66 | 1.23 | 2.00 | 2.64 | 3.33 |
| c_t | 330504 | 3.48 | 1.93 | 2.50 | 3.33 | 4.25 |
| ct_t | 330504 | 0.72 | 0.31 | 0.60 | 0.80 | 1.00 |
| dc_c | 330504 | 0.16 | 0.13 | 0.07 | 0.15 | 0.24 |
| dc_t | 330504 | 0.61 | 0.60 | 0.20 | 0.50 | 0.88 |
| cp_c | 330504 | 0.28 | 0.28 | 0.12 | 0.23 | 0.36 |
| cp_t | 330504 | 0.91 | 0.88 | 0.38 | 0.75 | 1.20 |
| t_s | 330504 | 0.74 | 0.32 | 0.59 | 0.80 | 1.00 |
| cn_c | 330504 | 1.33 | 0.63 | 1.00 | 1.27 | 1.58 |
| cn_t | 330504 | 4.57 | 2.80 | 3.13 | 4.29 | 5.67 |
| vp_t | 330504 | 1.78 | 0.89 | 1.40 | 1.75 | 2.13 |

Table S6: Descriptive Statistics: Control Group (U.S. NES Authors)

| Name | Company | Country | Issuing date |
|---|---|---|---|
| Chatgpt-3.5 | OpenAI | United States | 2022.11.30 |
| Chatgpt-4 | OpenAI | United States | 2023.3.14 |
| Chatgpt-4o | OpenAI | United States | 2024.5.13 |
| Chatgpt-o1 | OpenAI | United States | 2024.9.13 |
| Deepseek-v3 | Deepseek | China | 2024.12.16 |
| Qianfan-8b | Baidu | China | 2025.4.24 |
| Llama-4-Scout | Meta | United States | 2025.4.6 |

Table S7: Large Language Model Specifications for Text-Polishing Experiments.

| Id | Text |
|---|---|
| prompt1 | I wrote an abstract for my research, please help me polish the text. Do not modify the meanings. The output should be the polished text only without any other explanation or any other words. |
| prompt2 | Please refine the language of this abstract to improve clarity and flow without altering the core meaning. The output should be the polished text only without any other explanation or any other words. |
| prompt3 | Improve the wording and structure of this abstract to enhance its coherence and precision, maintaining the original message. The output should be the polished text only without any other explanation or any other words. |

Table S8. Instructional Prompts for LLM Text-Polishing Experiments.

| Variables | Word Length | | | | | |
|---|---|---|---|---|---|---|
| | (1) | (2) | (3) | (4) | (5) | (6) |
| *CN×PostGPT* | 0.080*** | 0.092*** | 0.085** | 0.094*** | 0.114*** | 0.097*** |
| | (0.009) | (0.007) | (0.011) | (0.007) | (0.007) | (0.008) |
| *Control Variables* | | Y | Y | Y | Y | Y |
| *Year FE* | Y | Y | Y | Y | Y | Y |
| *JCR Quartile FE* | | | Y | | | |
| *Journal FE* | | | | Y | | Y |
| *Author FE* | | | | | Y | Y |
| *Number of Journals* | | | | 30918 | | 30918 |
| *Number of Authors* | | | | | 1326001 | 1326001 |
| *Number of ob.* | 2801218 | 2801218 | 2578604 | 2801218 | 2801218 | 2801218 |
| $R^2$ | 0.001 | 0.003 | 0.017 | 0.088 | 0.686 | 0.707 |

Table S9. Multi-Model DID Estimates of ChatGPT Effects on NNES Word Length. *** p<0.001, ** p<0.01, * p<0.05.

| Variables | Word Length | | | |
|---|---|---|---|---|
| JIF Quartile | Q1 | Q2 | Q3 | Q4 |
| | (1) | (2) | (3) | (4) |
| *CN×PostGPT* | 0.085*** | 0.154*** | 0.119** | 0.090* |
| | (0.009) | (0.011) | (0.016) | (0.031) |
| *Control Variables* | Y | Y | Y | Y |
| *Year FE* | Y | Y | Y | Y |
| *Journal FE* | Y | Y | Y | Y |
| *Number of Journals* | 17302 | 7561 | 4364 | 1578 |
| *Number of ob.* | 2063662 | 344102 | 130665 | 40175 |
| $R^2$ | 0.06 | 0.236 | 0.484 | 0.224 |

Table S10. Heterogeneous Effects by Journal Tier (JIF Quartiles). *** p<0.001, ** p<0.01, * p<0.05.

| Variables | Word Length | | | |
|---|---|---|---|---|
| Domain | Life | Physical | Health | Social |
| | (1) | (2) | (3) | (4) |
| *CN×PostGPT* | 0.134** | 0.110*** | 0.074** | 0.150*** |
| | (0.018) | (0.009) | (0.016) | (0.015) |
| *Control Variables* | Y | Y | Y | Y |
| *Year FE* | Y | Y | Y | Y |
| *Journal FE* | Y | Y | Y | Y |
| *Number of Journals* | 15779 | 20545 | 18125 | 21643 |
| *Number of ob.* | 648266 | 1802917 | 674921 | 466354 |
| $R^2$ | 0.091 | 0.138 | 0.08 | 0.279 |

Table S11. Heterogeneous Effects by Domain. *** p<0.001, ** p<0.01, * p<0.05.

| Text | Detect Method | To be classified as | | Total |
|---|---|---|---|---|
| | | AI | Human | |
| A. | Original Text | | | |
| Original Abstract | detectllm | 17 | 83 | 100 |
| | fastdetectgpt | 2 | 98 | 100 |
| | loglikelihood | 18 | 82 | 100 |
| | logrank | 14 | 86 | 100 |
| B. | Polished Text | | | |
| Deepseek Polished | detectllm | 18 | 82 | 100 |
| | fastdetectgpt | 1 | 99 | 100 |
| | loglikelihood | 22 | 78 | 100 |
| | logrank | 15 | 85 | 100 |
| Chatgpt Polished | detectllm | 17 | 83 | 100 |
| | fastdetectgpt | 3 | 97 | 100 |
| | loglikelihood | 23 | 77 | 100 |
| | logrank | 14 | 86 | 100 |
| Llama Polished | detectllm | 25 | 75 | 100 |
| | fastdetectgpt | 3 | 97 | 100 |
| | loglikelihood | 28 | 72 | 100 |
| | logrank | 23 | 77 | 100 |
| Qianfan Polished | detectllm | 13 | 87 | 100 |
| | fastdetectgpt | 0 | 100 | 100 |
| | loglikelihood | 17 | 83 | 100 |
| | logrank | 9 | 91 | 100 |

Table S12: Failure of AI Detection Tools on Scientific Abstracts: False Positive and False Negative Rates. Performance evaluation across 500 texts (100 human + 400 AI-polished) using four detection methods

| Variables | $\Delta$Production | |
|---|---|---|
| | (1) lexical | (2) syntactic |
| *Interaction term* | 0.0013*** | 0.0003 |
| | (0.000) | (0.000) |
| *$\Delta$Complexity* | -0.0015*** | -0.0004 |
| | (0.000) | (0.000) |
| *NNES* | -0.0847*** | -0.0831*** |
| | (0.005) | (0.005) |
| *Control Variables* | Y | Y |
| *Number of ob.* | 477865 | 477865 |
| $R^2$ | 0.018 | 0.018 |

Table S13. Baseline regression results: The effect of changes in language complexity on the publication output differences between NNES and NES authors. *** p<0.001, ** p<0.01, * p<0.05.

| Variables | $\Delta$Production Rank | |
|---|---|---|
| | (1) lexical | (2) syntactic |
| *Interaction term* | -0.0105** | -0.0022 |
| | (0.003) | (0.003) |
| *$\Delta$Complexity* | 0.0085** | 0.0033 |
| | (0.003) | (0.003) |
| *NNES* | 0.2713*** | 0.2511*** |
| | (0.048) | (0.047) |
| *Control Variables* | Y | Y |
| *Number of ob.* | 477865 | 477865 |
| $R^2$ | 0.006 | 0.006 |

Table S14. Robustness check: Regression results using changes in publication rank as the dependent variable. *** p<0.001, ** p<0.01, * p<0.05.